%% file: main.tex
\documentclass[10pt,twocolumn,letterpaper]{article}

\usepackage[accsupp]{axessibility}
\usepackage[pagenumbers]{cvpr} %

\input{preamble}

\definecolor{cvprblue}{rgb}{0.21,0.49,0.74}
\usepackage[pagebackref,breaklinks,colorlinks,allcolors=cvprblue]{hyperref}

\def\paperID{15874} %
\def\confName{CVPR}
\def\confYear{2025}

\title{Edge-SD-SR: Low Latency and Parameter Efficient On-device Super-Resolution with Stable Diffusion via Bidirectional Conditioning}

\newcommand{\cmark}{\ding{51}}%
\newcommand{\xmark}{\ding{55}}%
\author{%
  Isma Hadji\textsuperscript{1*} \quad
  Mehdi Noroozi\textsuperscript{1*} \quad
  Victor Escorcia\textsuperscript{1} \quad
  Anestis Zaganidis\textsuperscript{1} \quad \\
  Brais Martinez\textsuperscript{1} \quad
  Georgios Tzimiropoulos\textsuperscript{1,2} \\
  {\normalsize $^1$Samsung AI Center, Cambridge, UK} \quad
  {\normalsize $^2$Queen Mary University of London, UK}
  \textsuperscript{*} \tt Equal contribution \quad
}

\begin{document}
\maketitle
\input{secs/abstract}

\input{secs/intro}

\input{secs/related}

\input{secs/approach}

\input{secs/experiments}

\input{secs/conclusion}
{
    \small
    \bibliographystyle{ieeenat_fullname}
    \bibliography{main}
}

\input{secs/X_suppl}

\end{document}

%% file: preamble.tex
\usepackage[utf8]{inputenc} %
\usepackage[T1]{fontenc}    %
\usepackage{url}            %
\usepackage{booktabs}       %
\usepackage{amsfonts}       %
\usepackage{nicefrac}       %
\usepackage{microtype}      %
\usepackage{xcolor}         %
\usepackage{amsmath}

\DeclareMathOperator*{\argmin}{arg\,min}
\DeclareMathOperator{\diag}{diag}

\usepackage{graphicx}
\usepackage{multirow}
\usepackage{xcolor}
\usepackage{comment}
\usepackage{pifont}%
\usepackage{multirow}
\usepackage{float}
\usepackage{makecell}
\usepackage{amsmath}
\usepackage{algpseudocode}
\usepackage{algorithm}
\usepackage{wrapfig}
\usepackage[
raggedright
]{sidecap}
\sidecaptionvpos{figure}{t}

%% file: secs/abstract.tex
\begin{abstract}

There has been immense progress recently in the visual quality of Stable Diffusion-based Super Resolution (SD-SR). However, deploying large diffusion models on computationally restricted devices such as mobile phones remains impractical due to the large model size and high latency. 
This is compounded for SR as it often operates at high res (\eg 4K$\times$3K). %
In this work, we introduce Edge-SD-SR, the \emph{first} parameter efficient and low latency diffusion model for image super-resolution. Edge-SD-SR consists of $\sim169$M parameters, including \textit{UNet, encoder and decoder}, and has a complexity of only $\sim142$ GFLOPs. 
To maintain a high visual quality on such low compute budget, we introduce a number of training strategies: (i) A novel conditioning mechanism on the low-resolution input, coined bidirectional conditioning, which tailors the SD model for the SR task. (ii) Joint training of the UNet and encoder, while decoupling the encodings of the HR and LR images and using a dedicated schedule. (iii) Finetuning the decoder using the UNet's output to directly tailor the decoder to the latents obtained at inference time.
Edge-SD-SR runs efficiently on device, e.g. it can upscale a $128\times128$ patch to $512\times 512$ in 38 msec while running on a Samsung S24 DSP, and of a $512\times512$ to $2,048\times2,048$ (requiring 25 model evaluations) in just $\sim 1.1$ sec. Furthermore, we show that Edge-SD-SR matches or even outperforms state-of-the-art SR approaches on the most established SR benchmarks.
\end{abstract}

%% file: secs/intro.tex
\section{Introduction}
\begin{figure}[t]
\begin{minipage}{0.32\columnwidth}
\includegraphics[width=\linewidth]{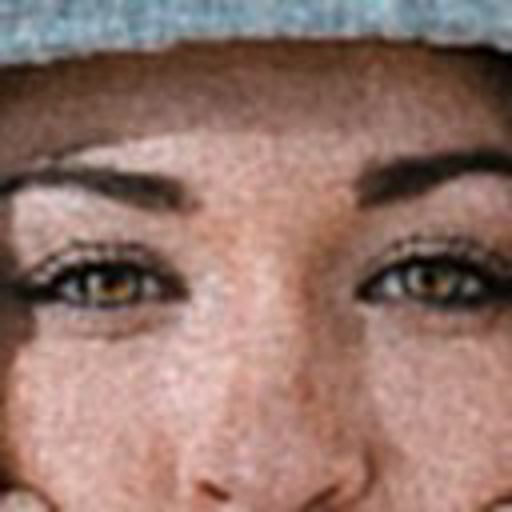}
\end{minipage}
\begin{minipage}{0.32\columnwidth}
\includegraphics[width=\linewidth]{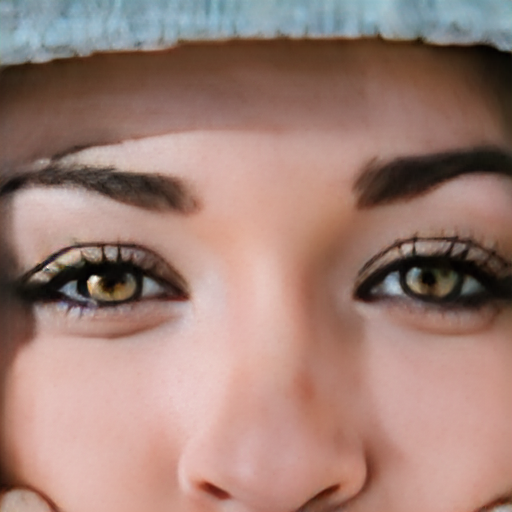}
\end{minipage}
\begin{minipage}{0.32\columnwidth}
\includegraphics[width=\linewidth]{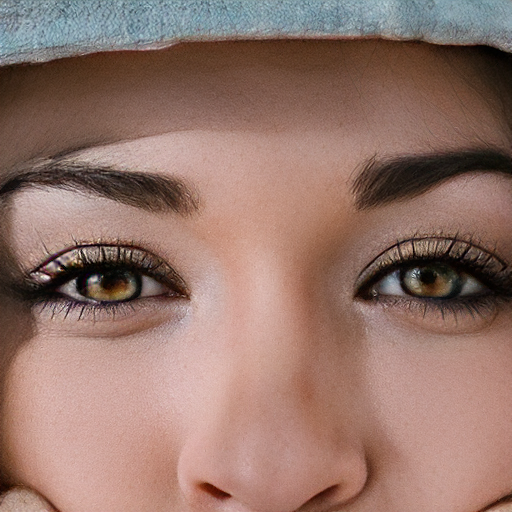}
\end{minipage}
\begin{minipage}{0.32\columnwidth}
\includegraphics[width=\linewidth]{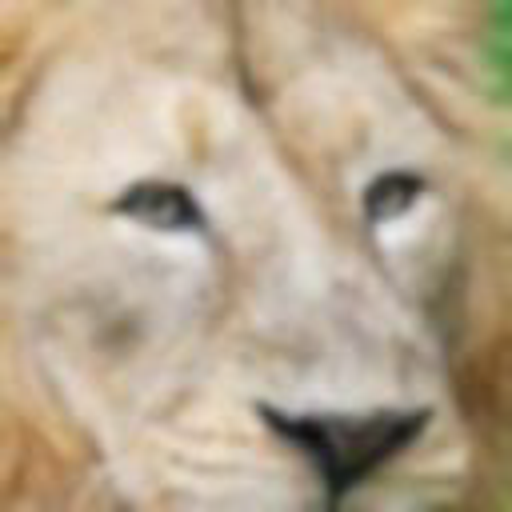}
\end{minipage}
\begin{minipage}{0.32\columnwidth}
\includegraphics[width=\linewidth]{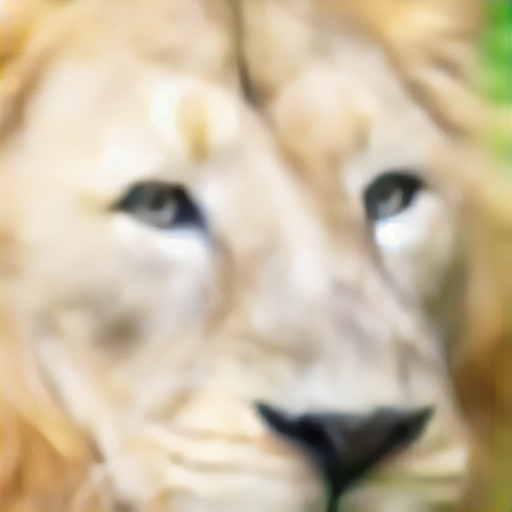}
\end{minipage}
\begin{minipage}{0.32\columnwidth}
\includegraphics[width=\linewidth]{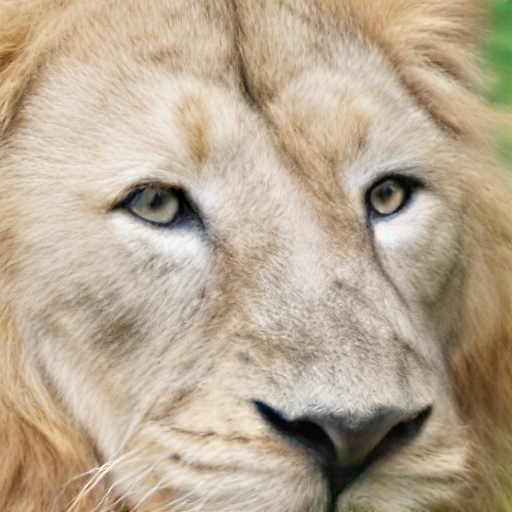}
\end{minipage}
\begin{minipage}{0.32\columnwidth}
\includegraphics[width=\linewidth]{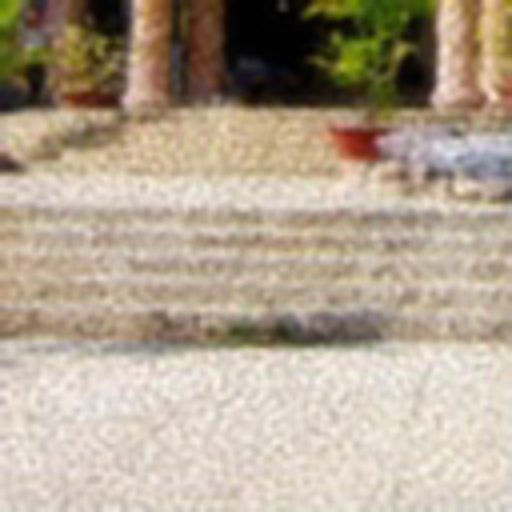}
\end{minipage}
\begin{minipage}{0.32\columnwidth}
\includegraphics[width=\linewidth]{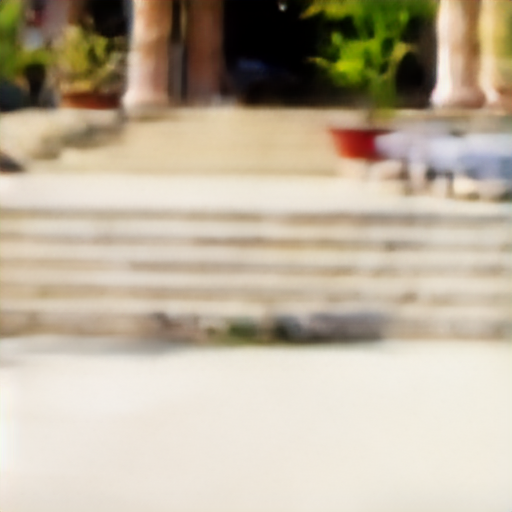}
\end{minipage}
\begin{minipage}{0.32\columnwidth}
\includegraphics[width=\linewidth]{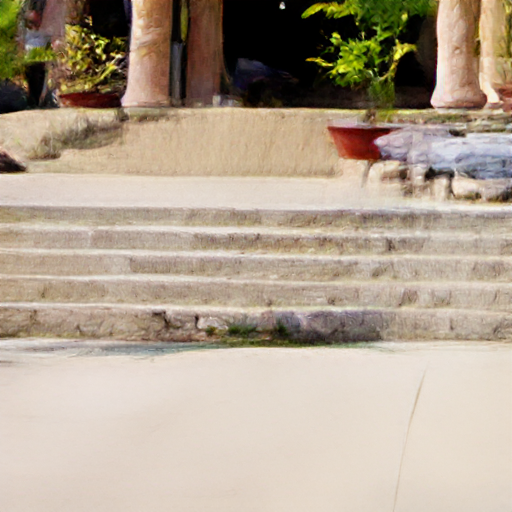}
\end{minipage}
\begin{minipage}{0.32\columnwidth}
\centering
(a)
\end{minipage}
\begin{minipage}{0.32\columnwidth}
\centering
~~(b)
\end{minipage}
\begin{minipage}{0.32\columnwidth}
\centering
~~(c)
\end{minipage}
    \caption{%
    Results of $4\times$ upsampling with (a) interpolation, (b) our architecture with the standard training approach   (c) our architecture with our training approach. Images are best seen in a display and zoomed in. Samples taken from DIV2K-RealESRGAN dataset.}
    \label{fig:teaser}
    \vspace{-15pt}
\end{figure}

We are interested in the application of Stable Diffusion~\cite{ldm_cvpr22,controlnet} to Super Resolution (\ie SD-SR), as it has been shown to produce steep improvements compared to previous SR approaches ~\cite{stablesr_arx23,yonos_arxiv24,resshift_neurips23}. Considering that SR is most widely used on mobile phones, where the vast majority of photos are taken nowadays, we focus on the very challenging task of creating a SD-SR model that can run on Edge devices (\ie Edge SD-SR). Deploying such models on computationally limited devices, such as mobile phones, entails addressing several practical challenges due to the large model size and high latency. These issues are exacerbated for the SR task, as it operates on high-resolution and typically requires multiple model runs.  
For example, 4K$\times$3K is a typical resolution on modern phones, while SD-SR typically generates $512\times512$ patches per run, \eg StableSR \cite{stablesr_arx23}. In this case, 88 model runs are required for tile-based generation. %
Specifically, %
to get a $4K\times3K$ image we split the corresponding $1000\times750$ LR image to $128\times128$ patches with $25\%$ overlap between patches. This yields 88 patches that the model has to process sequentially. %
Such a model runs in $\sim900$ seconds on a high-end GPU (A100), making on-device deployment unfeasible. 
In this work, we aim to address both model size and latency while maintaining the model's visual quality.

Improving diffusion models' efficiency has been a rather popular topic in recent years, typically focusing on reducing the number of denoising steps at inference~\cite{ddim,dpm,dpm++,song2023consistency,yonos_arxiv24}, or reducing the model size \cite{snapfusion_li2023,mobilediffusion}. However, to date, deploying a fast (1-step) \emph {and} light SD-SR model for real-time scenarios on edge devices remains an open area of exploration. 
In our work, we adopt prior art when possible, and focus on bridging the remaining gap. In particular, we 1) follow prior work \cite{snapfusion_li2023} and adopt a pruning and re-training strategy for the UNet (although a more aggressive pruning level is required in our case), 2) use compact variants of the encoder and decoder, which do not boast a large proportion of the parameters, but account for a significant portion of the FLOPs and latency, 3) follow \cite{yonos_arxiv24} to train a 1-step model, which includes a decoder fine-tuning strategy. Despite all these efforts to leverage existing work, the resulting light ($\sim169$M parameter) and efficient ($\sim142$ GFLOPS) model suffers from a large drop in image quality.  as illustrated in Fig.~\ref{fig:teaser}. Sec.~\ref{sec:main_ablation} provides further quantitative evaluations of the gap between existing technology and our goal. %

While it is tempting to attribute such performance drop to the extreme reduction of model capacity (\eg 850M to 158M params for the UNet) and step count (down to 1 step), we show that \textit{in the case of super-resolution}, 
such a drop can be largely mitigated through careful improvements to the optimization strategy. We hypothesize that SR requires less capacity than general text-to-image generation as it generates from an LR image instead of random noise.

Specifically, we introduce a new conditioning mechanism that takes the low-resolution image into account during \emph{both} the noising and denoising processes, which we coin bidirectional conditioning. This conditioning provides a better training signal, which allows us to train a very compact diffusion model to a high level of quality. In addition, we extend the training strategy to jointly train a lightweight encoder directly using the diffusion loss, thanks to an asymmetric encoder design and a modified noise sampling schedule. Finally, we combine our approach with the training strategy proposed in previous work \cite{yonos_arxiv24} targeting a 1-step SD-SR model, which also allows for directly training a diffusion-aware lightweight decoder tailored to the task of SR.

\begin{figure*}[t]
\begin{minipage}{0.19\textwidth}
\includegraphics[width=\linewidth]{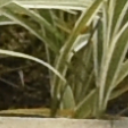}
\end{minipage}
\begin{minipage}{0.19\textwidth}
\includegraphics[width=\linewidth]{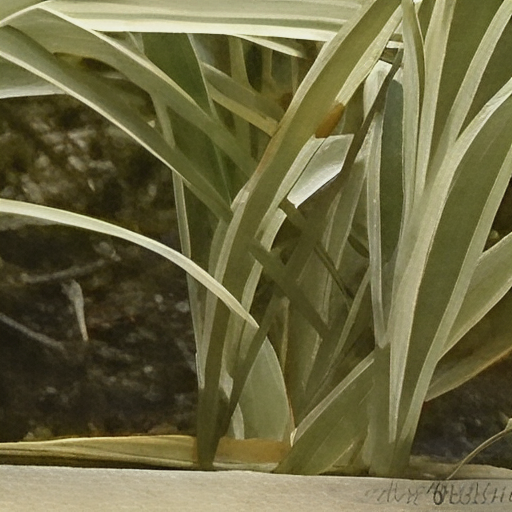}
\end{minipage}
\begin{minipage}{0.19\textwidth}
\includegraphics[width=\linewidth]{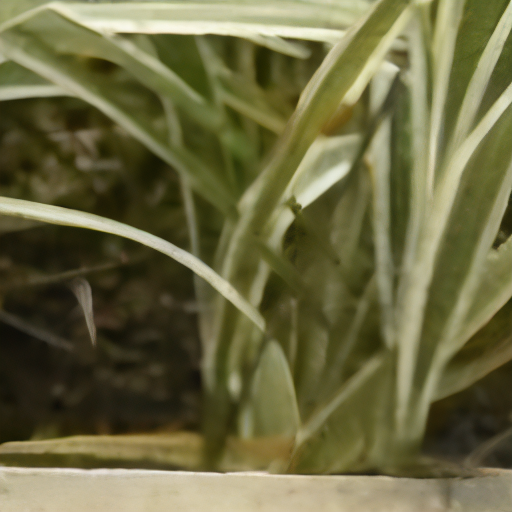}
\end{minipage}
\begin{minipage}{0.19\textwidth}
\includegraphics[width=\linewidth]{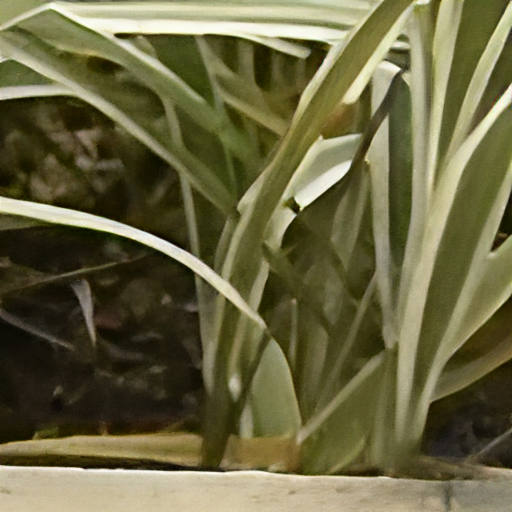}
\end{minipage}
\begin{minipage}{0.19\textwidth}
\includegraphics[width=\linewidth]{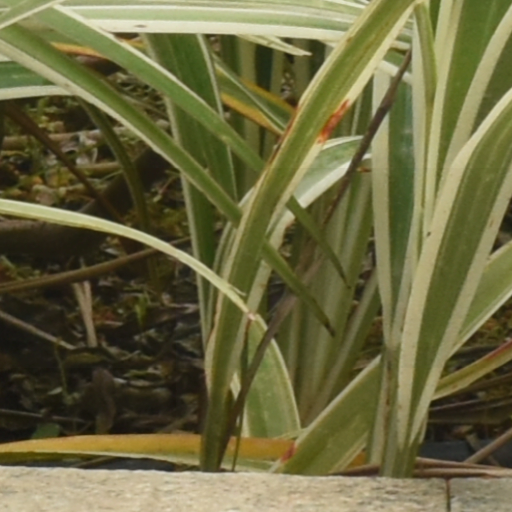}
\end{minipage}

\begin{minipage}{0.19\textwidth}
\includegraphics[width=\linewidth]{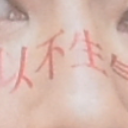}
\end{minipage}
\begin{minipage}{0.19\textwidth}
\includegraphics[width=\linewidth]{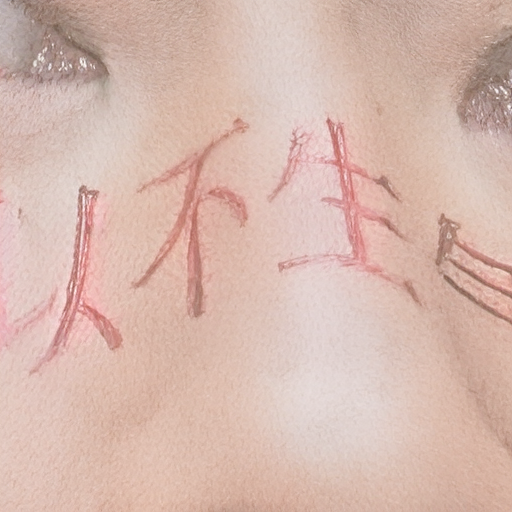}
\end{minipage}
\begin{minipage}{0.19\textwidth}
\includegraphics[width=\linewidth]{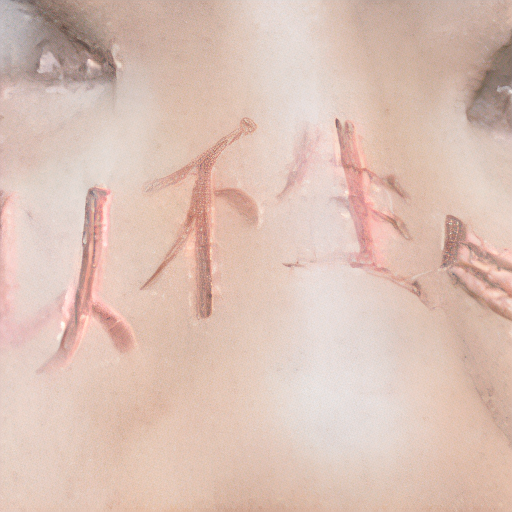}
\end{minipage}
\begin{minipage}{0.19\textwidth}
\includegraphics[width=\linewidth]{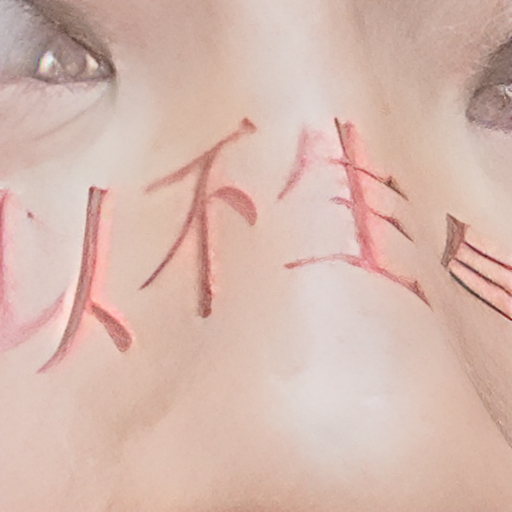}
\end{minipage}
\begin{minipage}{0.19\textwidth}
\includegraphics[width=\linewidth]{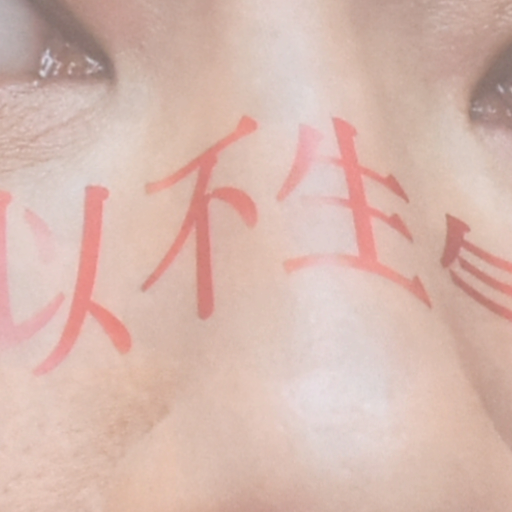}
\end{minipage}

\begin{minipage}{0.19\textwidth}
\includegraphics[width=\linewidth]{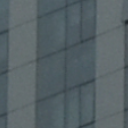}
\end{minipage}
\begin{minipage}{0.19\textwidth}
\includegraphics[width=\linewidth]{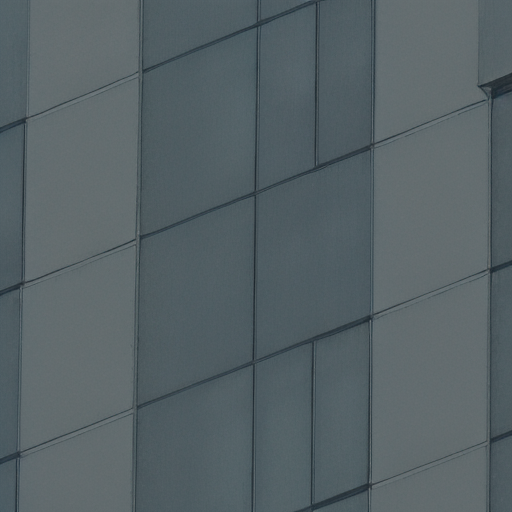}
\end{minipage}
\begin{minipage}{0.19\textwidth}
\includegraphics[width=\linewidth]{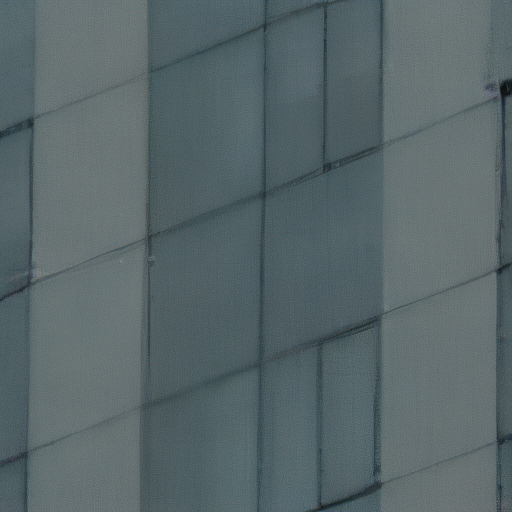}
\end{minipage}
\begin{minipage}{0.19\textwidth}
\includegraphics[width=\linewidth]{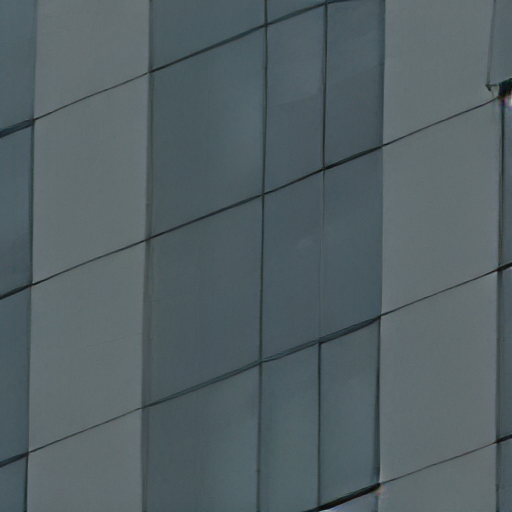}
\end{minipage}
\begin{minipage}{0.19\textwidth}
\includegraphics[width=\linewidth]{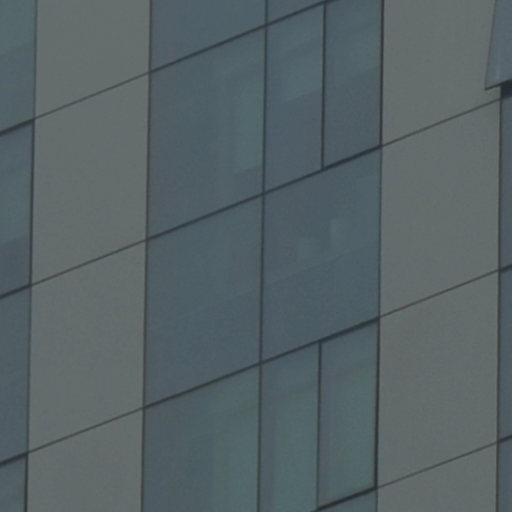}
\end{minipage}

\begin{minipage}{0.19\textwidth}
\includegraphics[width=\linewidth]{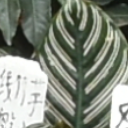}
\end{minipage}
\begin{minipage}{0.19\textwidth}
\includegraphics[width=\linewidth]{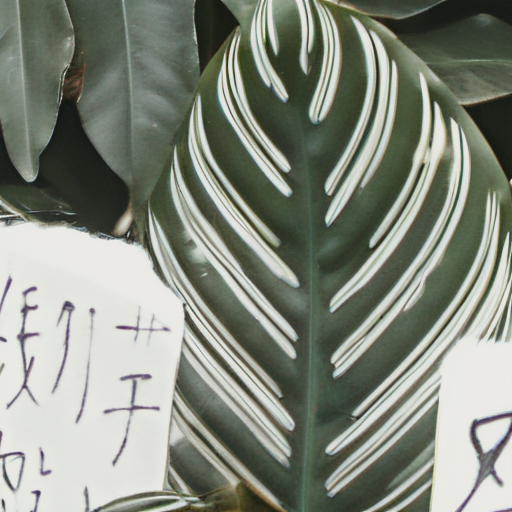}
\end{minipage}
\begin{minipage}{0.19\textwidth}
\includegraphics[width=\linewidth]{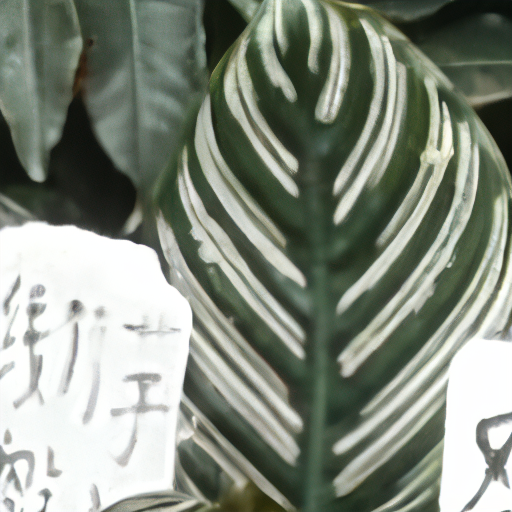}
\end{minipage}
\begin{minipage}{0.19\textwidth}
\includegraphics[width=\linewidth]{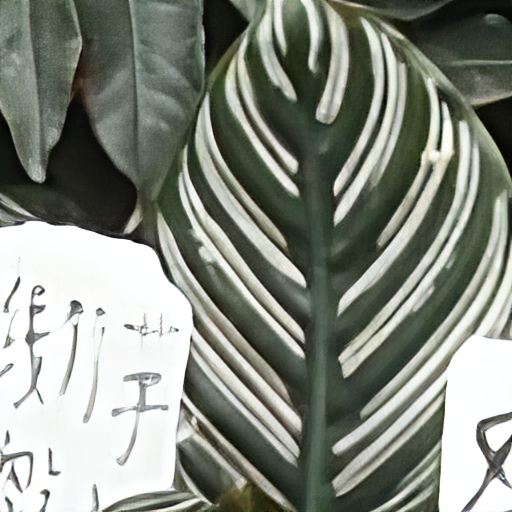}
\end{minipage}
\begin{minipage}{0.19\textwidth}
\includegraphics[width=\linewidth]{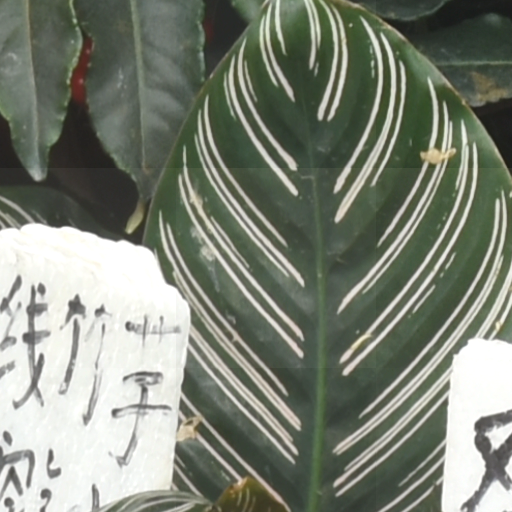}
\end{minipage}
\begin{minipage}{0.19\textwidth}

\end{minipage}

\begin{minipage}{0.19\textwidth}
\centering
(a)
\end{minipage}
\begin{minipage}{0.19\textwidth}
\centering
(b)
\end{minipage}
\begin{minipage}{0.19\textwidth}
\centering
(c)
\end{minipage}
\begin{minipage}{0.19\textwidth}
\centering
(d)
\end{minipage}
\begin{minipage}{0.19\textwidth}
\centering
(e)
\end{minipage}
\caption{Qualitative results. (a) input LR, (b) StableSR 200-steps (c) Resshift 15-Steps (d) Ours 1-step (e) ground truth. Samples are taken from DRealSR dataset. The images are best seen in a display and zoomed in. ResShift and StableSR include $173M$ and $1043M$ parameters and both require $> 2,000$ GFLOPs to process a $128\times128$ patch, respectively. Our model includes only 169M parameters and requires only 142 GFLOPs. Despite being hugely cost and parameter-efficient, our method works competitively with the larger baselines.}
\label{fig:qualititavie}
\vspace{-15pt}
\end{figure*}

To summarize, our contributions are threefold. 
I) We extend previous works to design a lightweight SD model tackling encoder+UNet+decoder. 
II) We propose bidirectional conditioning to effectively train a lightweight UNet for the SR task without compromising quality. 
III) We design a new strategy to jointly train the UNet and an efficient encoder tailored to model the LR image. 
The lightweight encoder and UNet are then combined with an efficient decoder trained using a method targeting 1-step SD-SR model, giving rise to a fully efficient SD-SR architecture.  Our final small model, coined Edge-SD-SR, has only $\sim169$M parameters and $\sim 142$ GFLOPS and can $4\times$ upscale a whole $512\times512$ image to $2,048\times2,048$ on-device (on a Samsung Galaxy S24) in just $\sim 1.1$ sec. Importantly, it achieves competitive results with large and slow state-of-the-art diffusion-based SR models (\eg ~\cite{stablesr_arx23, resshift_neurips23}) on established SR benchmarks (\eg. DIV2K, RealSR, DRealSR) as shown in Fig.~\ref{fig:qualititavie}.

%% file: secs/related.tex
\section{Related work} \label{sec:related}

\noindent \textbf{Image SR:} Image super-resolution involves enhancing a low-resolution (LR) image to reconstruct a high-resolution (HR) version of it. In this work, we are focusing on blind SR, where the image degradation parameters used to create the LR image are not assumed to be known. To model the LR-HR pair we follow the well-established RealESRGAN degradation pipeline \cite{realesrgan_iccvw21} which has shown satisfying generalization to real-world LR images. For a complete review of this heavily researched topic please see~\cite{wang2020deep}. 

\noindent \textbf{Diffusion for SR:} we focus SR based on diffusion models \cite{ldm_cvpr22}, which have shown remarkable capabilities, albeit at the cost of high compute cost, which we address in this work. The impressive performance of SD for generation tasks resulted in early interest in the SR community. Early work~\cite{sr3} adopted SD for SR in some constrained settings like face and class-constrained SR. General formulations for in-the-wild images were later proposed, \eg~\cite{sr3+}. Often works leverage image priors of SD models for SR \cite{codi_arxiv23,stablesr_arx23,diffbir,seesr}. For example, \cite{stablesr_arx23} exploited these priors by using adapters on top of a frozen SD model, achieving remarkable visual quality. More recent work \cite{yonos_arxiv24} proposed scale distillation and produced state-of-the-art results with only one inference step. However, these methods build on top of SD1.5 \cite{ldm_cvpr22}, which is highly parameter and latency-inefficient for on-device deployment. We consider instead the orthogonal problem of training a compact model effectively for SD-SR. Notably, our method is compatible with scale distillation, and we leverage it to train a one-step model.

A key component in our method is a new technique for conditioning the SD on the LR image, which we term bidirectional conditioning. The works most related to ours regarding this specific contribution are InDI~\cite{InDI} and ResShift~\cite{resshift_neurips23}. As we discuss in Sec.~\ref{ssec:bidirectional_conditioning}, InDI is not a diffusion model. They relax the degradation process to allow a direct transition from LR to HR. However, a direct translation of their formulation to diffusion models reduces to a special case of our formulation where the noise variance is not conditioned on the LR image. Interestingly, this direct translation is very similar to ResShift~\cite{resshift_neurips23}. Instead, we propose a more general formulation where the noise variance is also conditioned on the LR image and empirically show that conditioning the noise variance plays a vital role in the final performance. Importantly, none of these methods study efficiency. Particularly, we further propose a training paradigm for the encoder training and show that we can effectively reduce encoder size for an on-device model. %

\noindent \textbf{Efficient Diffusion:} Diffusion models are known for generating high-quality results at the cost of lower efficiency and therefore several works propose to tackle the efficiency aspect of diffusion models. There are two predominant approaches for making diffusion models more efficient. The first line of work focuses on reducing the number of \emph{inference steps} by proposing more effective sampling strategies \cite{ddim,dpm,dpm++}, or distillation techniques \cite{salimans2022progressive, adversarial_arxiv23,song2023consistency, yonos_arxiv24}, while still relying on large models. 
The second line of work focuses on making the SD model \emph{lightweight}~\cite{snapfusion_li2023, mobilediffusion}, but still use multiple steps for inference. 

We leverage the above know-how when possible. We use \cite{yonos_arxiv24} to achieve one single inference step for SD-SR, and follow the recipe of~\cite{snapfusion_li2023} to build a lightweight UNet, although a much higher parameter reduction rate is necessary for our purpose. Our method also requires a more lightweight auto-encoder.
Overall, there is no prior work that \textit{both} reduces inference steps \textit{and} uses a very lightweight model for the whole architecture (encoder+U-Net+decoder). We combine pre-existing works to this end, and show that novel training strategies are required to achieve high SR quality when using such a compact model.

%% file: secs/approach.tex
\section{Method}\label{sec:approach}

\begin{figure*}[t!]
\centering
\includegraphics[width=0.99\textwidth]{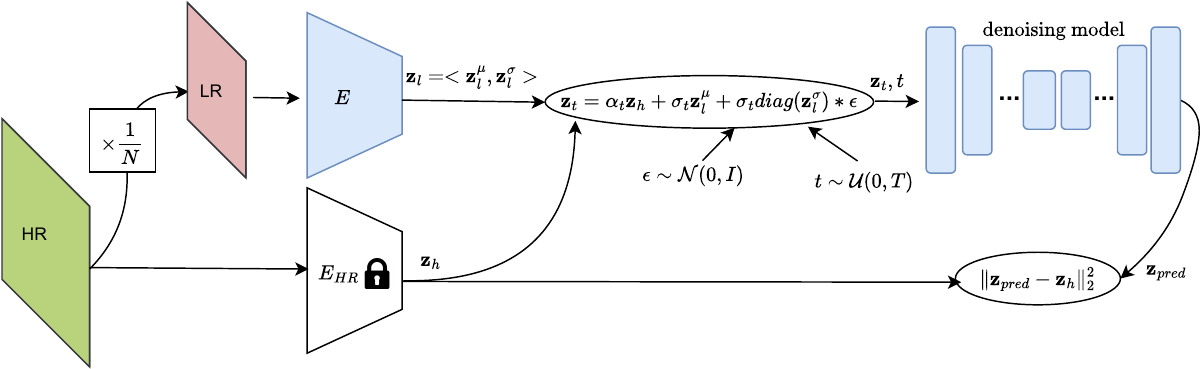}
\caption{\textbf{Overview of bidirectional conditioning for training}. We use different encoders for HR and LR, where $E_{HR}$ is frozen. We use the LR embeddings to condition the Gaussian mean and variance. The U-Net takes only $\mathbf{z}_t$ as input and is trained jointly with the encoder.}
\label{fig:bid_cond}
\end{figure*}

Our goal is to train an efficient SD-SR model that can readily run on Edge devices without compromising visual quality. Sec.~\ref{ssec:preliminaries} introduces the basics of SR-SD.
We then introduce our new bidirectional conditioning strategy, which is designed to better fit the SR task, in Sec.~\ref{ssec:bidirectional_conditioning}. 
We further devise a mechanism to train an efficient encoder in Sec.~\ref{ssec:enc-training}. In Sec.~\ref{ssec:dec-training} we discuss how the proposed strategies can be combined with previous work and expediting SD-SR with 1-step inference \cite{yonos_arxiv24} and training an efficient decoder. In Sec.~\ref{ssec:arch-size} we describe the adopted efficient architecture which we build based on previous works findings \cite{snapfusion_li2023,mobilediffusion} and extend to also squeeze the auto-encoder size. Finally, we present implementation details in Sec.~\ref{sec:implementation}.

\subsection{Stable diffusion for super-resolution}
\label{ssec:preliminaries}

We begin with preliminaries describing the standard SD-SR approach. Given a dataset of pairs of high-resolution (HR) and low-resolution (LR) images, $(\mathbf{x}_h, \mathbf{x}_l) \sim p(\mathbf{x}_h, \mathbf{x}_l)$, the objective of super-resolution (SR) is to estimate the probability distribution $p(\mathbf{x}_h|\mathbf{x}_l)$. Given that stable diffusion operates on the latent space of an encoder $\mathcal{E}$. The HR and LR images are first projected into latent space to yield $\mathbf{z}_h = \mathcal{E}(\mathbf{x}_h), \mathbf{z}_l = \mathcal{E}(\mathbf{x}_l)$, respectively.

During the training of SD-SR models, the HR image is progressively perturbed in a Markovian process, $q(\mathbf{z}_t|\mathbf{z}_h)$, through the addition of Gaussian noise conditioned on a schedule. The schedule is controlled by parameters $\alpha_t,\sigma_t$ such that the log signal-to-noise ratio, $\lambda_t = \log [{\alpha_t^2/\sigma_t^2}]$, decreases with $t$ monotonically. More specifically:

\begin{align}
q(\mathbf{z}_t|\mathbf{z}_h) = \mathcal{N}(\mathbf{z}_t;\alpha_t \mathbf{z}_h, \sigma_t \mathbf{I}), \mathbf{z}=\{ \mathbf{z}_t | t \in [0,1] \}
\label{eq:zt_unidirectional}
\end{align}

Note that the above forward noising process is the same as for the text-to-image case, except that it is not conditioned on an input caption but instead, it incorporates a conditioning mechanism on the LR image.

In particular, the LR image is given to the backward denoising function as an additional input; typically through concatenation with the noisy input. We can estimate the backward denoising process $p(\mathbf{z}_h|\mathbf{z}_t,\mathbf{z}_l)$ using a neural network, $\mathcal{Z}_\theta$, via a weighted mean square error loss:

\begin{align}
\argmin_\theta~ \mathbb{E}_{\epsilon,t}[\omega(\lambda_t) ||\mathcal{Z}_\theta(\mathbf{z}_t,\mathbf{z}_l,\lambda_t) - \mathbf{z}_h||_2^2]
\label{eq:standad_loss}
\end{align}

\noindent over uniformly sampled timesteps $t \in [0,1]$, and $
\mathbf{z}_t = \alpha_t \mathbf{z}_h +  \sigma_t\epsilon; \epsilon \sim \mathcal{N}(0,I).$

Starting from $\mathbf{z}_1 \sim \mathcal{N}(0,I)$, the inference process involves a series of sequential calls, i.e. steps, of $\mathcal{Z}_\theta$ where the quality of the generated image improves progressively with the number of steps. We use the widely used DDIM sampler in this paper \cite{ddim},

We call this mechanism unidirectional because the forward process \textbf{is not conditioned on the LR image}. We hypothesize that a direct conditioning on the low-res image of \textit{both} the forward and backward steps would provide a better signal to the training process. This, in turn, would put less pressure on the model, which otherwise needs to infer the LR information from the concatenated signal in the backward process. Ultimately, this would allow for reducing the model capacity. %
In the following, we introduce our bidirectional conditioning method to tackle this problem.

\subsection{Bidirectional conditioning}
\label{ssec:bidirectional_conditioning}

Our aim here is to modify the conditioning mechanism so that the forward process is also explicitly conditioned on the low-res image. To this end, we modify the Markovian Gaussian forward process in Eq.~\ref{eq:zt_unidirectional} of the diffusion to:

\begin{align}
q(\mathbf{z}_t|\mathbf{z}_h,\mathbf{z}_l) = \mathcal{N}(\mathbf{z}_t;\alpha_t \mathbf{z}_h + \sigma_t \mathbf{z}^\mu_l, \sigma_t\diag(\mathbf{z}^\sigma_l)) %
\label{eq:bidirectional}
\end{align}
 
\noindent where $\mathbf{z}_l=<\mathbf{z}^\mu_l,\mathbf{z}^\sigma_l>, \mathbf{z}_h$ are the embeddings of the LR and HR image, respectively, and $\mathbf{z}_h,\mathbf{z}^\mu_l,\mathbf{z}^\sigma_l$ are of the same dimensions. %
During training, the model learns to reverse this diffusion process progressively 
to generate the super-resolved image, \emph{without} needing $\mathbf{z}_l$ as an additional explicit input, given that it is now directly included in $\mathbf{z}_t$.  Specifically, we estimate the backward denoising process $p(\mathbf{z}_h|\mathbf{z}_t,\mathbf{z}_l)$ using a network, $\mathcal{Z}_\theta$, via a weighted mean square error loss:

\begin{align}
\argmin_\theta~ \mathbb{E}_{\epsilon,t}[\omega(\lambda_t) ||\mathcal{Z}_\theta(\mathbf{z}_t,\lambda_t) - \mathbf{z}_h||_2^2]
\label{eq:bidirectional_loss}
\end{align}

over uniformly sampled timesteps $t \in [0,1]$, and:

\begin{align}
\mathbf{z}_t =  \text{ }& \alpha_t \mathbf{z}_h +  \sigma_t  \mathbf{z}^\mu_l + \sigma_t \mathbf{z}^\sigma_l*\epsilon =  \alpha_t \mathbf{z}_h +  \sigma_t \hat\epsilon; \nonumber
\\  &  \epsilon \sim \mathcal{N}(0,I), \hat\epsilon \sim \mathcal{N}(\mathbf{z}^\mu_l,\diag(\mathbf{z}^\sigma_l))
\label{eq:zt}
\end{align}
\newline
\textbf{Sampling.} 
A direct benefit of the proposed formulation is that the resulting model predicts the HR image, at inference time, starting directly from the LR image instead of noise. We start from $ \mathbf{z}_1 \sim \mathcal{N}(\mathbf{z}^\mu_l,\diag(\mathbf{z}^\sigma_l))$, and call the denoising function progressively. More precisely, we use DDIM sampling \cite{ddim} as follows:
\begin{align}
z_{t-1} = & \sqrt{\alpha_{t-1}} \cdot  \mathcal{Z}_\theta(z_t) + \sqrt{1-\alpha_{t-1}-\sigma_t^2} \cdot  \mathcal{Z}_\theta(z_t) + 
 \sigma_t \hat{\epsilon}
\end{align}
Where $\hat\epsilon$ is sampled from the same $\mathcal{N}(\mathbf{z}^\mu_l,\diag(\mathbf{z}^\sigma_l))$ distribution at each step. In contrast, the unidirectional conditioning approach learns to generate the corresponding HR image starting from $\mathbf{z}_1 \sim \mathcal{N}(0,I)$ and progressively improves the result while having access to $\mathbf{z}_l$ in each iteration through concatenation, rather than the proposed more direct approach. Concretely, this means that our model has more direct access to the LR information rather than requiring the model to infer the information from a concatenated signal. Having this information allows for reducing the model size while maintaining quality, as shown in Figures~\ref{fig:teaser},~\ref{fig:qualititavie} and Sec.~\ref{sec:experiments}. %

\paragraph{Impact of noise variance conditioning.}

In the proposed bidirectional conditioning, we condition on the LR image, while taking into account both the mean and \emph{variance} as explicitly outlined in equations \eqref{eq:bidirectional} and \eqref{eq:zt}. Here, we want to highlight the importance of taking variance into account. To this end, we refer to recent work~\cite{InDI}, which proposed to learn a direct transition between LR and HR image, while assuming no knowledge of the forward degradation process. 

Our formulation is in line with them as we also learn a direct transition between the LR and HR image. However, we keep the degradation process the same as diffusion models, while conditioning the sampling Gaussian on the LR input, \ie we convert $q(\mathbf{z}_t|\mathbf{z}_h)$ to $q(\mathbf{z}_t|\mathbf{z}_h,\mathbf{z}_l$). Here, we show that a direct translation of their formulation to the stable diffusion framework, results in the following forward process:

\begin{align}
\mathbf{z}_t = \text{ } & \alpha_t \mathbf{z}_h +  \sigma_t  \mathbf{z}^\mu_l + \sigma_t\epsilon =  \alpha_t \mathbf{z}_h +  \sigma_t \tilde\epsilon; \nonumber \\  
&\epsilon \sim \mathcal{N}(0,I),  \quad \tilde\epsilon \sim \mathcal{N}(\mathbf{z}^\mu_l,I)
\label{eq:zt_indi}
\end{align}

This degradation process, which is also similar to the degradation process used by ResShift~\cite{resshift_neurips23}, is restricted compared to ours because only the mean of the sampling Gaussian distribution is conditioned on the LR image. We conjecture that taking the LR image into account for the noise variance is of paramount importance for the SR task to recover details in the resulting HR image. In fact, this contrasting point between our proposed bidirectional conditioning method and previous work \cite{InDI,resshift_neurips23} is more clear when analyzing the inference process. Rather than starting from random noise at inference time, previous work \emph{only} shifts the mean through the LR input image, thereby providing less details about the LR image to the denoising process compared to our approach. As a result, in these works, the LR image is still provided to the denoising model, $\mathcal{Z}_\theta$, explicitly (\ie through concatenation with $\mathbf{z}_t$) at each step. As such, these approaches rely on the model's capacity to infer those details from this additional input. By conditioning the variance as well, we provide more complete information about the LR image in our conditioning term, thereby not requiring the LR image as an additional input to the model.

In Sec.~\ref{sec:main_ablation}, we study the impact of our conditioning and the noise variance, \ie equations \eqref{eq:zt} vs \eqref{eq:zt_indi}, on our target efficient network. We observe significant superiority in our more general formulation where the noise variance is also conditioned on the LR input.

\subsection{Encoder Training}
\label{ssec:enc-training}

The SR methods based on stable diffusion use the same encoder for the HR and LR images. This encoder is pre-trained on the HR images~\cite{stablesr_arx23, resshift_neurips23, codi_arxiv23}. To cope with the domain shift between the LR and HR images, an \emph{additional} adaptor network is usually used ~\cite{stablesr_arx23, codi_arxiv23}, resulting in computational overhead. Alternatively, we propose to train the encoder jointly with the UNet for the SR task. This approach allows us to not only avoid computational overhead but also to reduce the encoder size drastically, where we need far fewer parameters to encode the LR image compared to the HR one. In particular, we use two encoders, $\mathcal{E}$ and $\mathcal{E}_{hr}$, for the LR and HR images, respectively. We initialize $\mathcal{E}_{hr}$ with the SD v1.5 text-to-image encoder and freeze it, %
while we initialize $\mathcal{E}$ randomly. We train $\mathcal{E}$ jointly with the UNet as illustrated in Fig.~\ref{fig:bid_cond}. Notably, the HR image encoder is only needed at training time, while we only need the efficient LR encoder at inference time.

Let us define $\mathcal{E}^\mu_\phi(x_l) = z^\mu_l$, and $\mathcal{E}^\sigma_\phi(x_l) = z^\sigma_l$, where $\mathcal{E}_\phi$ is the LR encoder network parametrized by $\phi$. We can train the encoder by adding its parameters to the optimization using the same equation ~\eqref{eq:bidirectional_loss}:

\begin{align}
\argmin_{\theta,\phi}~ \mathbb{E}_{\epsilon,t}[\omega(\lambda_t) || & \mathcal{Z}_\theta(\alpha_t \mathbf{z}_h + \sigma_t \mathcal{E}^\mu_\phi(x_l) + \nonumber \\  
& \sigma_t \mathcal{E}^\sigma_\phi(x_l)*\epsilon,\lambda_t) - \mathbf{z}_h||_2^2] 
\label{eq:bidirectional_loss_withenc}
\end{align}

\noindent \textbf{Preventing the degenerate solution}. The encoder is time-agnostic and, as a result, the vanilla training of the objective function in equation~\eqref{eq:bidirectional_loss_withenc} results in a degenerate solution for the average $t$. To prevent this, we perform a stop gradient technique for $t<1$ for the encoder. That is, when $t=1$, we optimize for both $\theta, \phi$, and we optimize only for $\theta$ when $t < 1$. Since $t$ is uniformly sampled at training time from a large number of discrete steps, e.g. 1000 steps, the solution above is prone to reduce to equation~\eqref{eq:bidirectional_loss} where only the denoising model is optimized. To address this issue, instead of sampling uniformly over  $t \in [0,1]$, we use a scheduler where we start more frequently for $t=1$ in the beginning, and we gradually drop the $t=1$ sampling rate until we converge to uniform sampling in the later training iterations.

The encoder training paradigm described above can be potentially applied to the unidirectional conditioning approach. However, as we show in Sec.~\ref{sec:main_ablation}, it is more effective when applied with our proposed bidirectional conditioning method, which we conjecture is thanks to the more direct use of the LR information.

\subsection{Decoder Training}\label{ssec:dec-training}
The proposed method allows us to effectively train a more efficient Encoder and UNet. We are however still left with a large decoder. As our goal is to fit the overall model on device, we need a method to also train a small decoder architecture. The naive solution is to swap the decoder architecture for a more efficient one and pretrain the auto-encoder following the same approach adopted in the original SD work \cite{ldm_cvpr22}. %
However, given that we are re-training \emph{both} the UNet and encoder with our approach, it is more judicious to finetune the decoder with the new encoder and UNet in place. This strategy allows the decoder to adapt to the new latent space yielding from our training which alters both the encoder and UNet. Unfortunately, given that the UNet requires several denoising steps to yield a good input for the decoder, this strategy incurs a non-trivial training cost, shall we use several denoising steps during training. We therefore need to use a method for reducing the number of denoising steps required by the UNet, to effectively train the decoder. Another benefit of reducing the number of inference steps is that it would directly address the speed requirement of our target Edge-SD-SR model.

The proposed approach can be readily combined with techniques targeting speeding up inference of the SD model, such as faster samplers \cite{ddim,dpm,dpm++}, progressive
distillation \cite{salimans2022progressive,song2023consistency} or scale distillation \cite{yonos_arxiv24}. In this work, we opt to use scale distillation as it was specifically designed for the SR task and demonstrated superior performance. Specifically, we first train a teacher model for a lower magnification factor (e.g. $\times2$) and progressively distill the teacher to a student model at the target scale factor of interest (e.g.$\times4$). We use our proposed joint training loss, see equation~\eqref{eq:bidirectional_loss_withenc}, in all cases.

The final step of our pipeline is to train the decoder on top of the efficient 1-step model. To this end, we use the same VQAE loss used to train the original auto-encoder in the original SD paper \cite{ldm_cvpr22}. This step concludes the training of the proposed Edge-SD-SR model, which can now be used with one 1 denoising step thanks to the combination with scale distillation approach \cite{yonos_arxiv24}.

\subsection{Efficient architecture}\label{ssec:arch-size}

In this work, we focus on devising a technique to train an Edge-SD-SR model \emph{without} sacrificing quality as described in Sections~\ref{ssec:bidirectional_conditioning} through~\ref{ssec:dec-training}. For the architectural choices of Edge-SD-SR, we mainly follow recent work (e.g. \cite{snapfusion_li2023,mobilediffusion}) to modify the architecture of the UNet used in SD 1.5 by (i) trimming the expansion ratio in the number of channels by half and (ii) focusing transformer blocks in lower resolution only. In addition, we propose to halve the number of base channels, as we found this approach to yield the biggest gain in terms of parameters and FLOPS. These modifications yield an efficient UNet, that has only $158M$ parameters and uses around $40$ GFLOPS to process a $128\times 128$ patch (vs. $860M$ parameters and $222$ GFLOPS for SD 1.5).

The auto-encoder used in SD 1.5 is also heavy with about $83M$ parameters and $1820$ GFLOPS. Therefore, we also reduce the size of the auto-encoder, following recent work (e.g. \cite{snapfusion_li2023,mobilediffusion}) by halving the number of channels and residual blocks. In addition, we completely remove one downsampling layer from the LR encoder, $\mathcal{E}$, such that it directly scales down the input image of size $128\times128$, by $4$. In contrast, the HR encoder, $\mathcal{E}_{hr}$, scales down HR images, of size $512\times512$, by $8$. Importantly, given the proposed decoupling of HR and LR encoder described in Sec.~\ref{ssec:enc-training}, the HR encoder can be kept frozen and is only required at training time, while we can deploy the efficient LR encoder at inference time. These design choices result in a lightweight auto-encoder with $14M$ parameters that require only $102$ GFLOPS to process a $128\times 128$ patch. 

Edge-SD-SR model is readily deployable on-device using standard quantization techniques. Using vanilla weights--int8 and activations--int16 (W8A16)
quantization, Edge-SD-SR can $4\times$ upscale a $128\times128$ patch on S24 NPU (QNN 2.19) in $\sim38$ msecs (encoder: 7.465 msecs, UNet: 8.340 msec, decoder: 22.453 msec). Moreover, to $4\times$ upscale a whole $512\times512$ image to $2,048\times2,048$ (25 model evaluations in total) this takes $\sim 1.1$ (including the tiling operations). Notably, we did not observe any quantitative or qualitative degradation by the quantization process. Improved latency could be achieved by further quantizing the activations.

\input{secs/implementation}

%% file: secs/implementation.tex
\subsection{Implementation details}\label{sec:implementation}
In addition to the architectural changes, aiming at an efficient architecture, described in Sec~\ref{ssec:arch-size}, we also follow recent work ~\cite{Emu} to increase the number of output channels of the encoder from $4$ to $8$ for improved performance. Consequently, we also increase the number of input channels of the UNet to $8$. For better performance with a 1-step model we adopt the scale distillation training paradigm \cite{yonos_arxiv24}. For optimization, we use the Adam optimizer and a fixed learning rate of $5e-5$ to first train the Encoder-UNet as described in Sec~\ref{ssec:enc-training} and train for 300 epochs. Finally, to train the decoder, we use the original VQAE loss \cite{ldm_cvpr22} and a learning rate of $1.8e-5$ and train using the Adam optimizer for 20 epochs.

%% file: secs/experiments.tex
\section{Experiments}
\label{sec:experiments}
\begin{table*}[t]
\resizebox{.99\textwidth}{!}{
    \centering
    \begin{tabular}{c|c|cccccc|c}
        \hline
         Datasets & Metrics & LDM & StableSR & YONOS & DiffBir & SeeSR & ResShift & Edge-SD-SR (Ours) \\
         \hline
        \multirow{5}{*}{\makecell{DIV2K Valid \\ RealESRGAN degradations} }
        & FID $\downarrow$ & 26.47 & \textcolor{blue}{24.44} & \textcolor{red}{21.86}&40.42 &31.93 &30.45&25.37\\
         & LPIPS $\downarrow$ & 0.2510 & 0.3114 & \textcolor{red}{0.2310}&0.427&0.3843&0.3076&\textcolor{blue}{0.249}\\
         & PSNR $\uparrow$ & 23.32 & 23.26 & \textcolor{red}{24.74}&20.94&21.19&\textcolor{blue}{24.62}&24.10\\
         & SSIM $\uparrow$ & 0.5762 &  0.5726 & \textcolor{red}{0.6428}&0.493&0.5386&\textcolor{blue}{0.6210}&0.617\\
         & MUSIQ $\uparrow$ & 62.27 & 65.92 & \textcolor{red}{70.30}&62.05&68.33&63.58&\textcolor{blue}{69.58} \\
         \hline
        \multirow{2}{*}{RealSR} & LPIPS $\downarrow$ & 0.3159 & 0.3002 & \textcolor{red}{0.2479}&0.3658&0.3009&0.3279&\textcolor{blue}{0.278}\\
        & MUSIQ $\uparrow$ & 58.90 & \textcolor{blue}{65.88} & 69.21&64.85&\textcolor{red}{69.77}&59.87&65.20\\
         \hline
         \multirow{2}{*}{DRealSR}  & LPIPS $\downarrow$ & 0.3379 & 0.3284 & \textcolor{red}{0.2721}&0.4599&0.3189&0.3870&\textcolor{blue}{0.292}\\
         & MUSIQ $\uparrow$ & 53.72 & \textcolor{blue}{58.51} & \textcolor{red}{66.26}&61.19&64.93&54.13&55.66\\
         \hline
         \multirow{1}{*}{DPED-iphone} & MUSIQ $\uparrow$ & 44.23 & 50.48 & \textcolor{blue}{59.45}&-&-&38.59&\textcolor{red}{60.09}\\
         \hline
         \hline
         - & \#~STEPS $\downarrow$ & 200 & 200 & \textbf{\textcolor{red}{1}}&50&50&4& \textbf{\textcolor{red}{1}} \\
         - & \#~GFLOPS $\downarrow$ & 9820 & 47403 & 2042& $>2000$ & $>2000$ &2651& \textbf{\textcolor{red}{142}} \\
        - & \#~Params (M) $\downarrow$ & 241 & 1063 & 960&1717&2524&173& \textbf{\textcolor{red}{169}} \\
         \hline
    \end{tabular}
    }
    \caption{\textbf{Comparison to stable diffusion-based baselines}. Results in \textcolor{red}{Red} and \textcolor{blue}{Blue} correspond to best and second best results respectively.}\label{tab:comp-to-sd}
    \vspace{-10pt}
\end{table*}

We now extensively evaluate our proposed method. We first summarize our experimental setup in Sec.~\ref{sec:exp-setup}. We then compare the lightweight model resulting from our proposed training strategy with state-of-the-art in Sec.~\ref{sec:sota_comparison}. Finally, we present an ablation study in Sec.~\ref{sec:main_ablation} to highlight the impact of the proposed bidirectional conditioning and training strategies on the final lightweight SD-based SR model.

\input{secs/experimental_setup}
\input{secs/experiments_SOTA_comparision}

\input{secs/experiments_design_choices}

%% file: secs/experimental_setup.tex
\subsection{Experimental setup}\label{sec:exp-setup}
\paragraph{Training datasets.} We follow previous work, \eg \cite{yonos_arxiv24,stablesr_arx23,femasr_mm22, realesrgan_iccvw21,zhang2021designing}, and train our model using a combination of DIV2K \cite{div}, DIV8K\cite{div8k}, Flickr2k \cite{flickr2k}, OST \cite{ost} and a subset of 10K images from FFHQ training set \cite{ffhq}. To generate the LR-HR pairs necessary for training, we use the Real-ESRGAN \cite{realesrgan_iccvw21} degradation pipeline. 

\paragraph{Testing datasets.} Similarly, at test time we adopt the same test setup used in more recent SR works, \eg \cite{yonos_arxiv24,stablesr_arx23}. Specifically, we use a synthetic dataset and three real datasets. For the synthetic dataset, we use a set of 3K LR-HR ($128\rightarrow512$) pairs synthesized from the DIV2K validation set using the Real-ESRGAN degradation pipeline. For the real datasets, we also follow previous work and use $128\times128$ center crops from the RealSR \cite{realsr_cvprw20}, DRealSR \cite{drealsr} and DPED-iphone \cite{dped} datasets.

\paragraph{Baselines.} We compare to a sample from recent stable diffusion-based SR methods, including the original LDM \cite{ldm_cvpr22}, StableSR \cite{stablesr_arx23}, Yonos-SR \cite{yonos_arxiv24} and ResShift \cite{resshift_neurips23}. %

\paragraph{Evaluation metrics.} We evaluate using standard perceptual and image quality metrics, including LPIPS\cite{lpips}, FID \cite{fid} (where applicable), and a no-reference image quality metric, MUSIQ \cite{musiq}. We also report PSNR and SSIM on the synthetic data for reference.

%% file: secs/experiments_SOTA_comparision.tex
\subsection{Comparison with state-of-the art}\label{sec:sota_comparison}

In this section, we evaluate the performance of our proposed Edge-SD-SR model in the standard real image super-resolution setting targeting $\times 4$ scale factor and compare it to state-of-the-art. As our main point of contribution is diffusion-based SR models, we focus on comparisons to SD-based models in Table~\ref{tab:comp-to-sd}. It is important to note here that this comparison is not fair by design as we operate on a challenging setting with \textbf{(1)} a very small model and \textbf{(2)} only consider results obtained with one denoising step. In contrast, each of the compared methods only targets one of these aspects that we tackle collectively in our work. All SD-SR methods compared here consume $> 2000$ GFLOPs, whereas our model requires $142$ GFLOPs. Specifically, the original LDM model \cite{ldm_cvpr22} uses a \emph{similarly sized U-Net} (but larger auto-encoder) and requires 200 denoising steps, yet we outperform this model in all datasets and metrics. StableSR \cite{stablesr_arx23} uses a much larger model and 200 denoising steps but focuses on \emph{encoder training} and once again we outperform this model in $7$ out of $10$ comparison points. YONOS-SR \cite{yonos_arxiv24} use a much larger model, but specifically optimize for \emph{inference speed} to obtain a model running in only 1 denoising step. This is a very challenging comparison point given that a similar strategy is adopted in our case to tackle inference speed but we do that with a very limited computational budget and as such we fall behind YONOS-SR in several comparison points. More closely related to ours is ResShift \cite{resshift_neurips23}, which also proposes a new conditioning strategy to rely more heavily on the LR image and uses only $4$ denoising steps. Notably, while we use a much smaller model size (considering the auto-encoder size as well), our proposed model outperforms ResShift in $8$ out of $10$ comparison points. In addition, we show quantitatively in Figure~\ref{fig:qualititavie} that our model yields superior or on-par results while being significantly smaller and faster than state-of-the-art baselines. Notably, we also outperform more recent SD-based SR work like DiffBir \cite{diffbir} and SeeSR \cite{seesr}. Collectively, these results support the effectiveness of the proposed approach to enable SD-SR models to readily run on-device.

%% file: secs/experiments_design_choices.tex
\subsection{Ablation study}\label{sec:main_ablation}
Here, we analyze the impact of each component that we proposed to yield our final Edge-SD-SR model. For this ablation we use the set of 3K LR-HR pairs synthesized from the DIV2K validation set to extensively study the impact of our contributions on a controlled large test set. We report FID and LPIPS scores, as we found these two metrics to align most with human assessment of the resulting HR images. 

\paragraph{Impact of bidirectional conditioning.}
We begin by evaluating the impact of the proposed bidirectional conditioning. To this end, we evaluate two variants of the proposed small model. Specifically, we train one model with the proposed bidirectional conditioning and another one with the standard unidirectional conditioning (\ie conditioning on the LR image via concatenation). Notably, given that the evaluated architecture uses a small encoder initialized from scratch, we have to train both the encoder and UNet in both cases. We train both models using the method described in Sec.~\ref{ssec:enc-training} and \emph{only} change the conditioning. We otherwise, keep all other training details identical. The results in the second row of Table~\ref{tab:ablations} speak decisively in favor of the proposed bidirectional conditioning, which outperforms the widely used unidirectional conditioning by a significant margin.
\vspace{-15pt}
\paragraph{Impact of noise variance conditioning.}\label{sec:compare_with_indi}
To study the impact of variance when conditioning the sampling Gaussian on the LR image, we set the sampling variance to identity, as described in equation~\eqref{eq:zt_indi}, and retrain our model. As we can see in the last row of Table~\ref{tab:ablations}, removing the variance from the conditioning, results in a significant drop in performance, thereby emphasizing the superiority of our formulation. 

\begin{table}[t]
    \centering
    \resizebox{0.99\columnwidth}{!}{
    \begin{tabular}{c|c|c|c}
         \toprule
         bi-directional & variance conditioning & FID $\downarrow$& LPIPS $\downarrow$ \\
         \hline
         \xmark & \xmark & 46.46 & 0.468\\
         \cmark & \cmark & \textbf{25.
         37} &\textbf{0.249}\\
         \cmark & \xmark & 39.13 & 0.313\\
         
         \bottomrule
    \end{tabular}
    }
    \caption{Impact of bidirectional conditioning.}
    \label{tab:ablations}
    \vspace{-10pt}
\end{table}

\begin{table}[t]
    \centering
    \resizebox{0.99\columnwidth}{!}{
    \begin{tabular}{c|c|c}
         \toprule
         Encoder and Decoder training method & FID $\downarrow$& LPIPS $\downarrow$ \\
         \hline
         Independent encoder training & 63.03 & 0.549\\
         Independent decoder training  & 52.78 & 0.401\\
         Edge-SD-SR & \textbf{25.
         37} &\textbf{0.249}\\
         \bottomrule
    \end{tabular}
    }
    \caption{Impact of encoder and decoder training.}
    \label{tab:enc-ablations}
   \vspace{-15pt} 
\end{table}

\paragraph{Impact of encoder training.}
While the results in Table~\ref{tab:ablations} implicitly highlight the importance of encoder training using our formulation vs. the standard unidirectional formulation, here we explicitly evaluate our joint encoder-UNet training strategy described in Sec.~\ref{ssec:enc-training}. To this end, we run an experiment where we first train the efficient LR encoder, $\mathcal{E}$, \emph{independently} of the UNet. The results in the first row of Table~\ref{tab:enc-ablations} highlight the paramount importance of adapting the LR images latents while training the UNet under our formulation because those latents are explicitly used to condition the response of the UNet.

\paragraph{Impact of decoder training.}
We finally, emphasize the importance of finetuning a UNet-aware small decoder. As mentioned in Sec.~\ref{ssec:dec-training}, and similar to the encoder case, a naive solution is to train an efficient decoder \emph{independently} of the diffusion model. Specifically, we replace the original big decoder with our efficient architecture and retrain the auto-encoder using the standard VQAE loss \cite{ldm_cvpr22}, while keeping the encoder frozen. Results in the second row of Table~\ref{tab:enc-ablations} show this is a sub-optimal solution leading to significantly worse results. In contrast, training a small decoder on top of the output of the UNet (keeping the encoder and UNet frozen) yields the best results. We argue that this strategy allows the decoder to adapt to the representation seen at inference time given that, in our formulation, both the encoder and UNet are trained. Notably, we use 1 denoising step in all cases given that we adopt scale distillation training.

\noindent Together, the results shown throughout this section, highlight the importance of tailoring the SD formulation to tackle the SR task. Our results show that by taking \emph{full} advantage of the LR image at training and inference time, we can effectively train a low computational budget model that can readily compete with larger baselines as shown in Table~\ref{tab:comp-to-sd}.

%% file: secs/conclusion.tex
\section{Conclusion}
In summary, we presented Edge-SD-SR, the \emph{first} SD-based SR solution that can readily run on device. While previous works tackle different aspects of making the diffusion models more efficient (\ie speed or size), we propose the first complete solution that tackles all aspects collectively. To enable training this model, we introduce a new conditioning technique that provides a better training signal for the diffusion model to more directly transition from LR to HR. Finally, we show that this method can readily be combined with techniques specifically targeting improving inference speed of SD-based SR to yield an efficient and fast model.

%% file: secs/X_suppl.tex
\clearpage
\appendix

\section{Additional quantitative results}
\begin{figure*}
\begin{minipage}{0.19\textwidth}
\includegraphics[width=\linewidth]{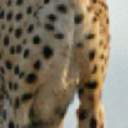}
\end{minipage}
\begin{minipage}{0.19\textwidth}
\includegraphics[width=\linewidth]{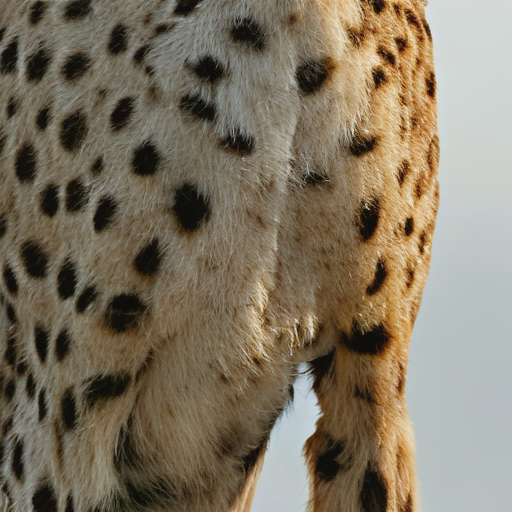}
\end{minipage}
\begin{minipage}{0.19\textwidth}
\includegraphics[width=\linewidth]{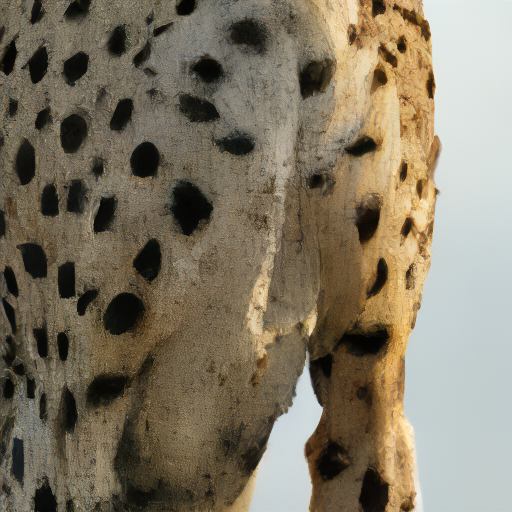}
\end{minipage}
\begin{minipage}{0.19\textwidth}
\includegraphics[width=\linewidth]{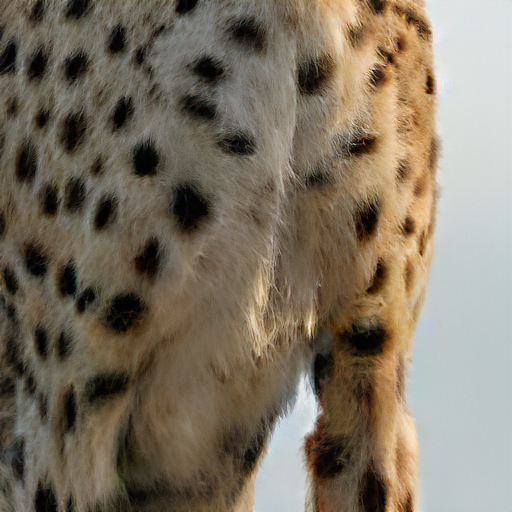}
\end{minipage}
\begin{minipage}{0.19\textwidth}
\includegraphics[width=\linewidth]{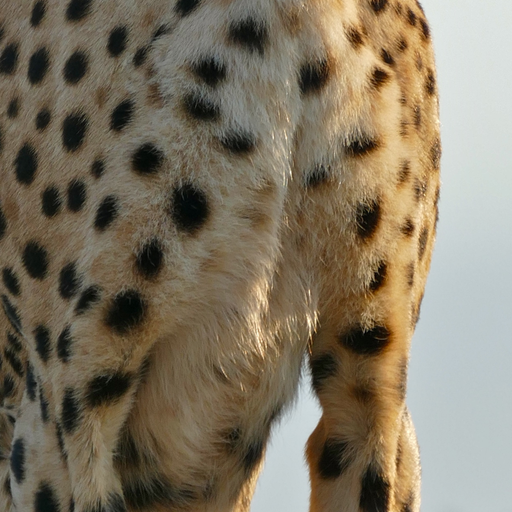}
\end{minipage}
\begin{minipage}{0.19\textwidth}
\includegraphics[width=\linewidth]{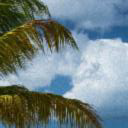}
\end{minipage}
\begin{minipage}{0.19\textwidth}
\includegraphics[width=\linewidth]{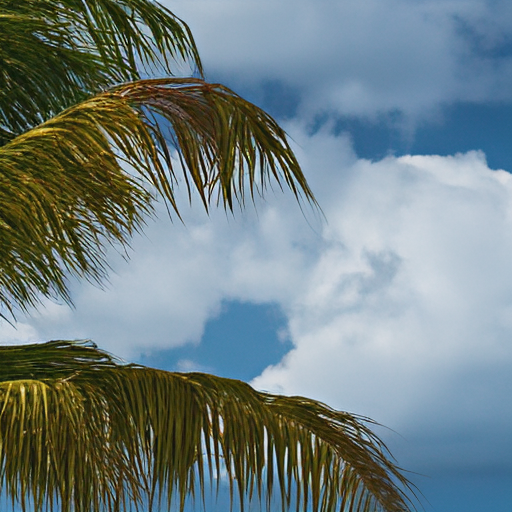}
\end{minipage}
\begin{minipage}{0.19\textwidth}
\includegraphics[width=\linewidth]{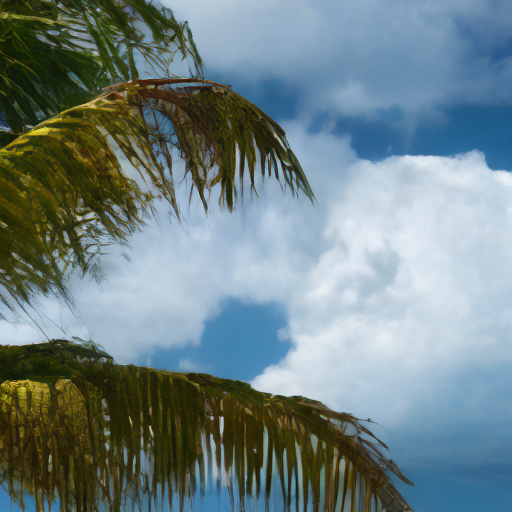}
\end{minipage}
\begin{minipage}{0.19\textwidth}
\includegraphics[width=\linewidth]{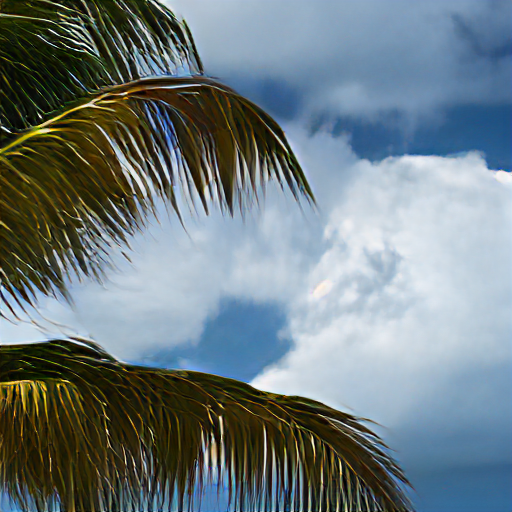}
\end{minipage}
\begin{minipage}{0.19\textwidth}
\includegraphics[width=\linewidth]{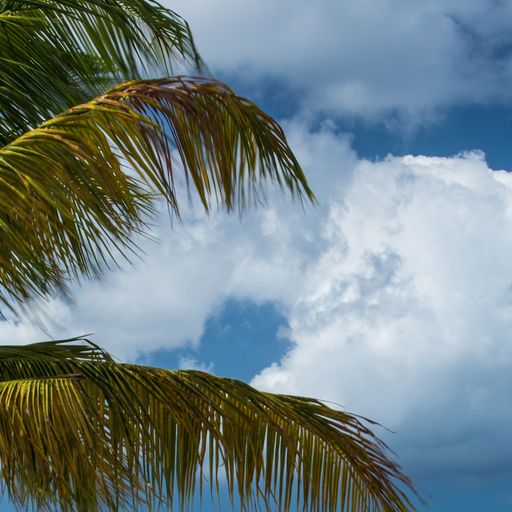}
\end{minipage}
\begin{minipage}{0.19\textwidth}
\includegraphics[width=\linewidth]{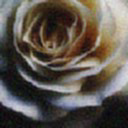}
\end{minipage}
\begin{minipage}{0.19\textwidth}
\includegraphics[width=\linewidth]{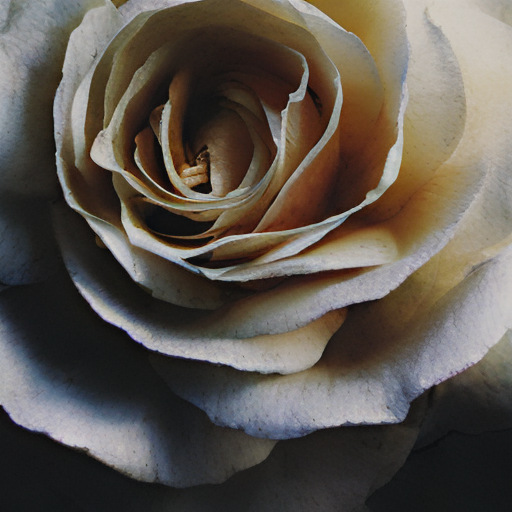}
\end{minipage}
\begin{minipage}{0.19\textwidth}
\includegraphics[width=\linewidth]{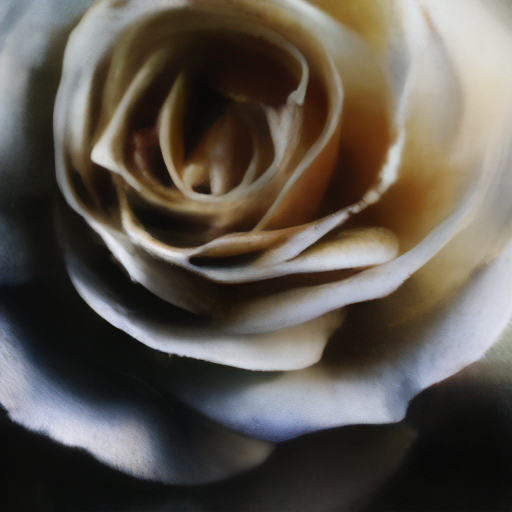}
\end{minipage}
\begin{minipage}{0.19\textwidth}
\includegraphics[width=\linewidth]{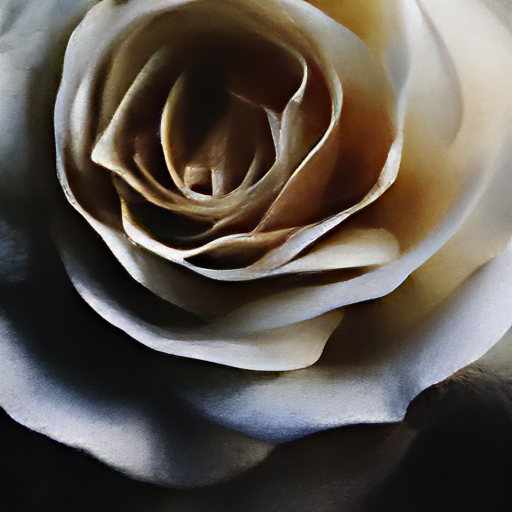}
\end{minipage}
\begin{minipage}{0.19\textwidth}
\includegraphics[width=\linewidth]{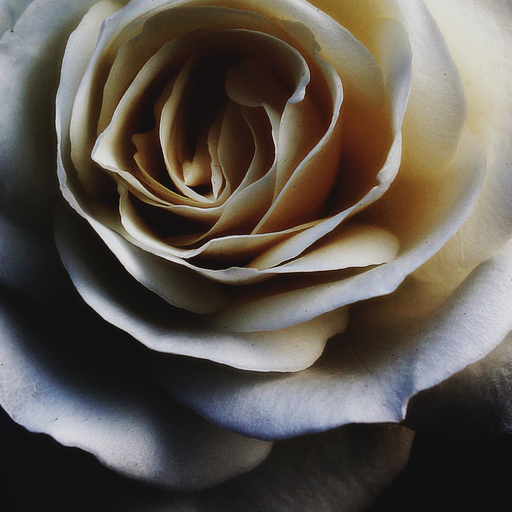}
\end{minipage}
\begin{minipage}{0.19\textwidth}
\includegraphics[width=\linewidth]{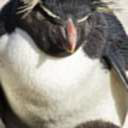}
\end{minipage}
\begin{minipage}{0.19\textwidth}
\includegraphics[width=\linewidth]{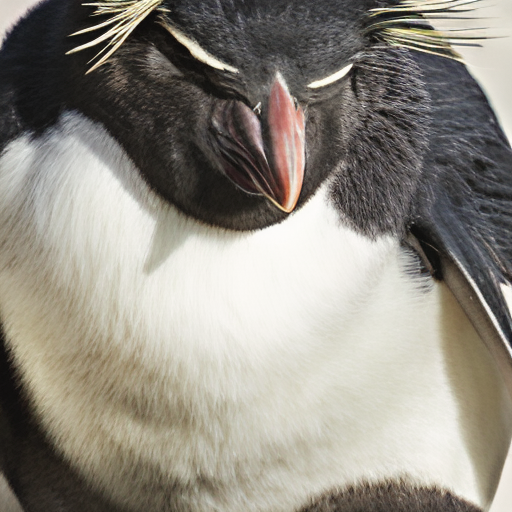}
\end{minipage}
\begin{minipage}{0.19\textwidth}
\includegraphics[width=\linewidth]{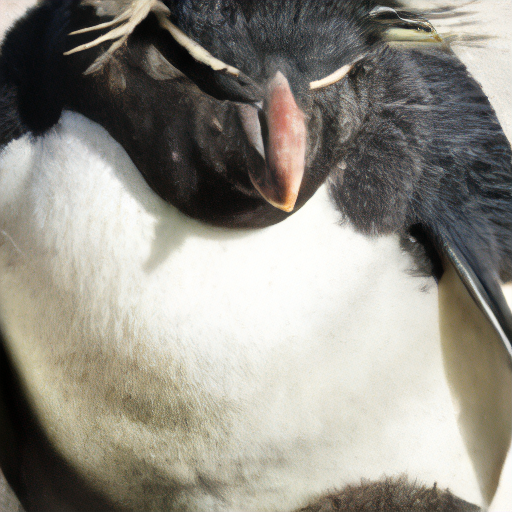}
\end{minipage}
\begin{minipage}{0.19\textwidth}
\includegraphics[width=\linewidth]{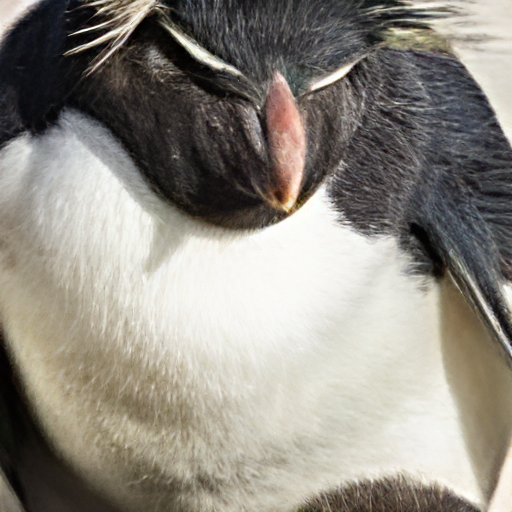}
\end{minipage}
\begin{minipage}{0.19\textwidth}
\includegraphics[width=\linewidth]{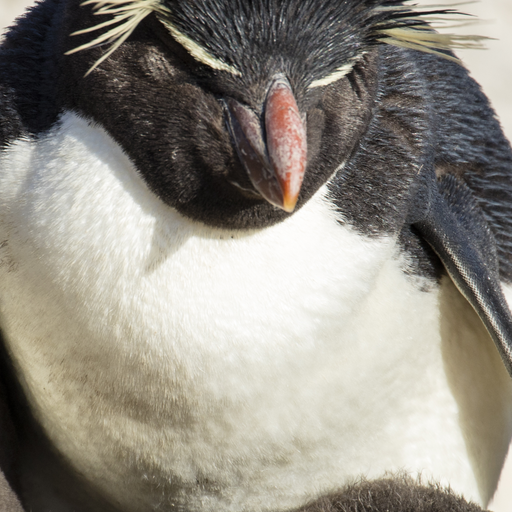}
\end{minipage}
\begin{minipage}{0.19\textwidth}
\centering
(a)
\end{minipage}
\begin{minipage}{0.19\textwidth}
\centering
(b)
\end{minipage}
\begin{minipage}{0.19\textwidth}
\centering
(c)
\end{minipage}
\begin{minipage}{0.19\textwidth}
\centering
(d)
\end{minipage}
\begin{minipage}{0.19\textwidth}
\centering
(e)
\end{minipage}
\caption{Qualitative results on samples from the DIV2K-RealESRGAN dataset. (a) input LR, (b) StableSR 200-steps (c) Resshift 15-Steps (d) Ours 1-step (e) ground truth. ResShift and StableSR include $173M$ and $1043M$ parameters and both require $> 2,000$ GFLOPs to process a $128\times128$ patch, resp. Ours includes only 169M parameters and requires 142 GFLOPs. Despite being hugely cost and parameter-efficient, our method works competitively with the larger baselines.}
\label{fig:sup-synth}
\end{figure*}

\begin{figure*}
\begin{minipage}{0.19\textwidth}
\includegraphics[width=\linewidth]{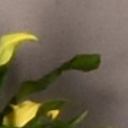}
\end{minipage}
\begin{minipage}{0.19\textwidth}
\includegraphics[width=\linewidth]{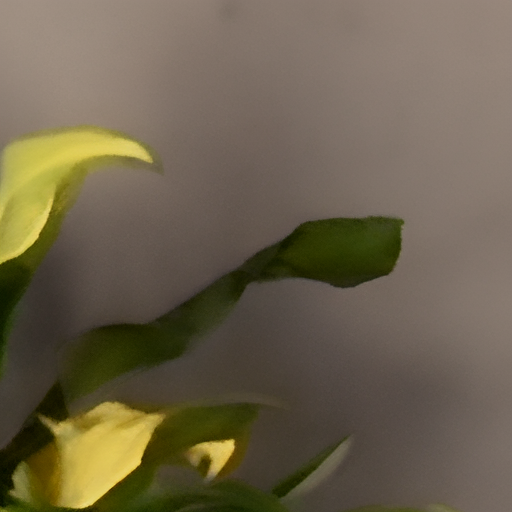}
\end{minipage}
\begin{minipage}{0.19\textwidth}
\includegraphics[width=\linewidth]{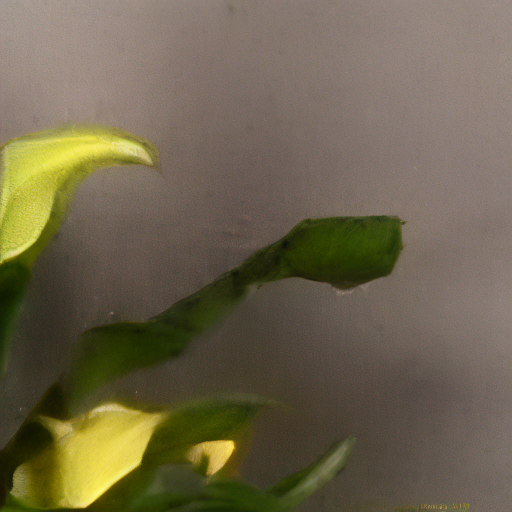}
\end{minipage}
\begin{minipage}{0.19\textwidth}
\includegraphics[width=\linewidth]{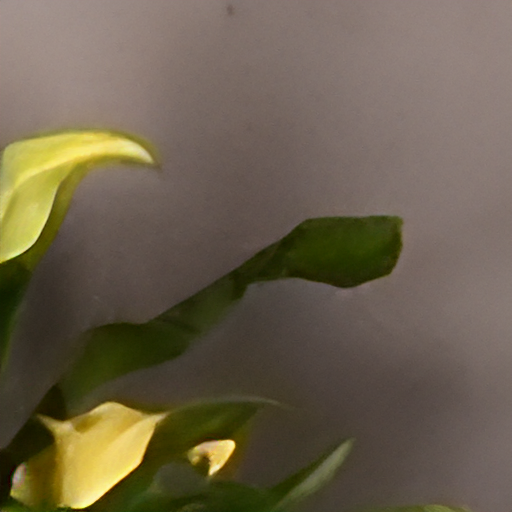}
\end{minipage}
\begin{minipage}{0.19\textwidth}
\includegraphics[width=\linewidth]{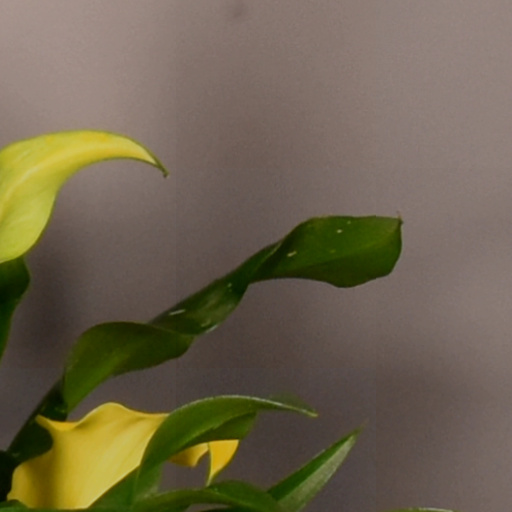}
\end{minipage}
\begin{minipage}{0.19\textwidth}
\includegraphics[width=\linewidth]{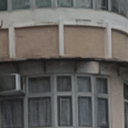}
\end{minipage}
\begin{minipage}{0.19\textwidth}
\includegraphics[width=\linewidth]{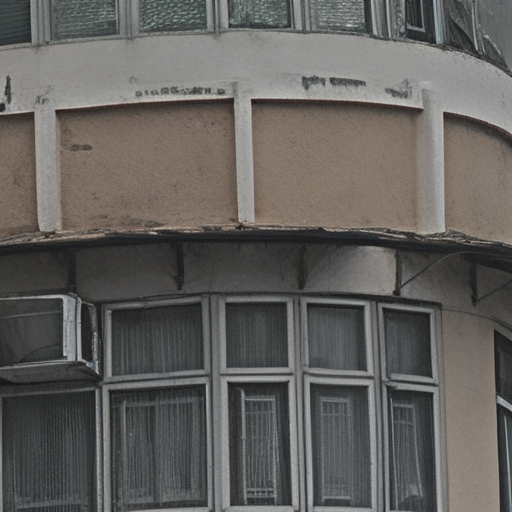}
\end{minipage}
\begin{minipage}{0.19\textwidth}
\includegraphics[width=\linewidth]{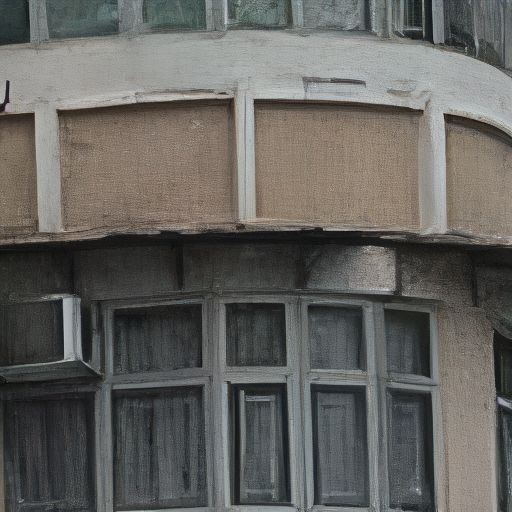}
\end{minipage}
\begin{minipage}{0.19\textwidth}
\includegraphics[width=\linewidth]{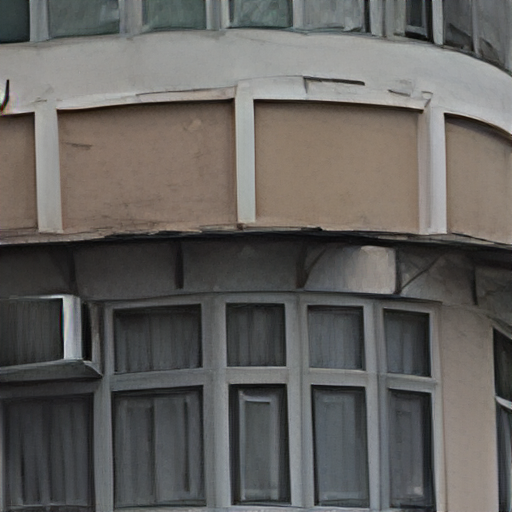}
\end{minipage}
\begin{minipage}{0.19\textwidth}
\includegraphics[width=\linewidth]{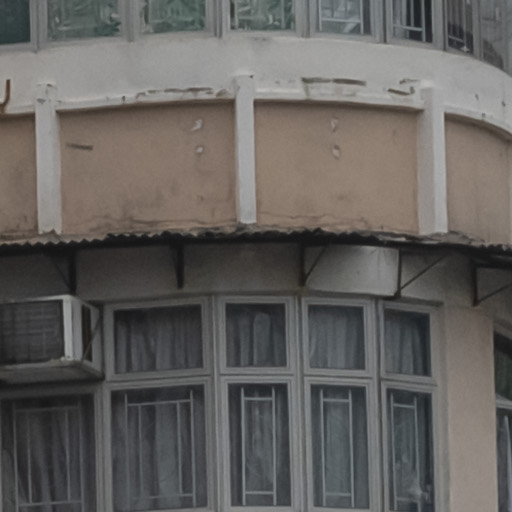}
\end{minipage}
\begin{minipage}{0.19\textwidth}
\includegraphics[width=\linewidth]{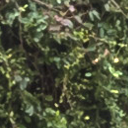}
\end{minipage}
\begin{minipage}{0.19\textwidth}
\includegraphics[width=\linewidth]{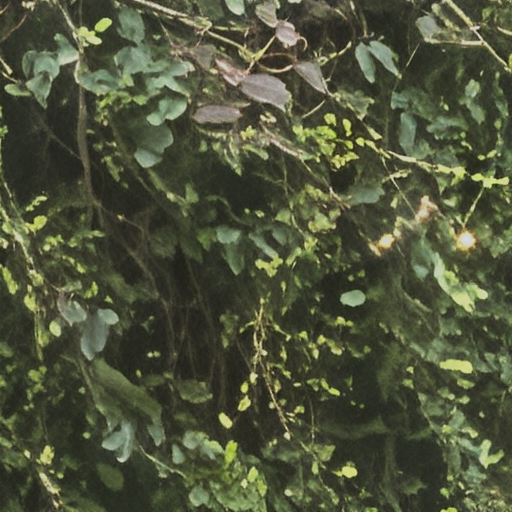}
\end{minipage}
\begin{minipage}{0.19\textwidth}
\includegraphics[width=\linewidth]{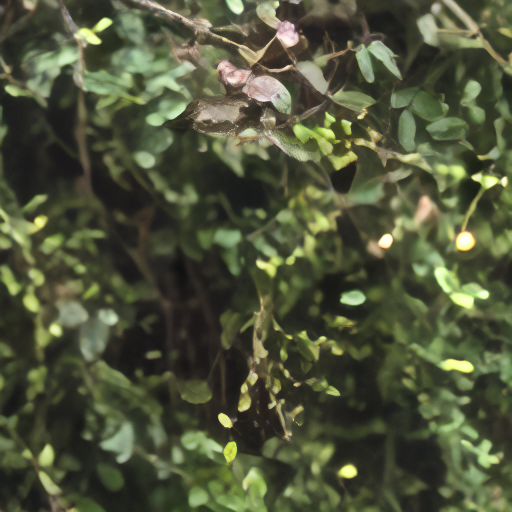}
\end{minipage}
\begin{minipage}{0.19\textwidth}
\includegraphics[width=\linewidth]{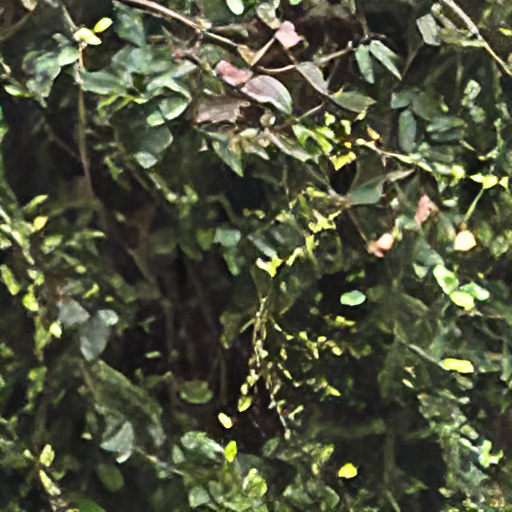}
\end{minipage}
\begin{minipage}{0.19\textwidth}
\includegraphics[width=\linewidth]{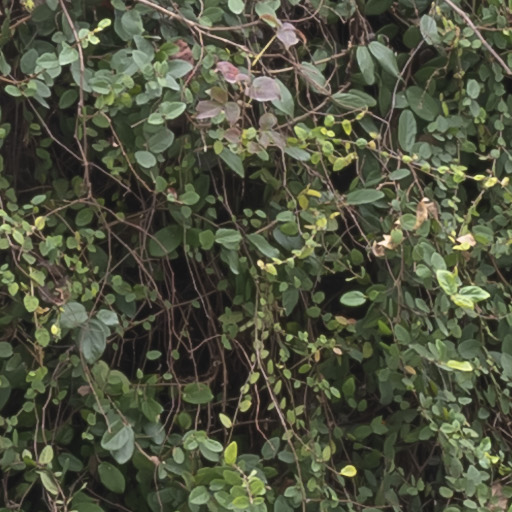}
\end{minipage}
\begin{minipage}{0.19\textwidth}
\includegraphics[width=\linewidth]{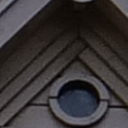}
\end{minipage}
\begin{minipage}{0.19\textwidth}
\includegraphics[width=\linewidth]{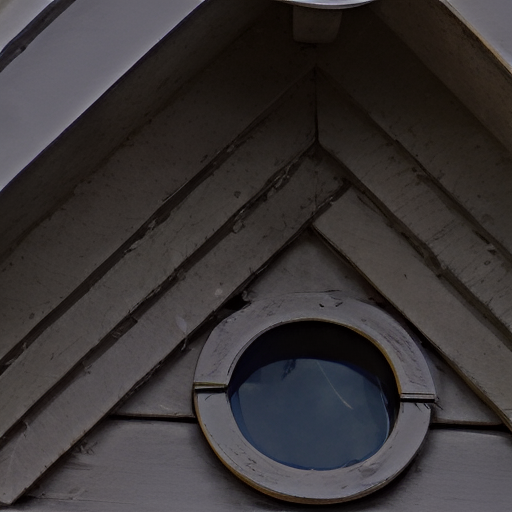}
\end{minipage}
\begin{minipage}{0.19\textwidth}
\includegraphics[width=\linewidth]{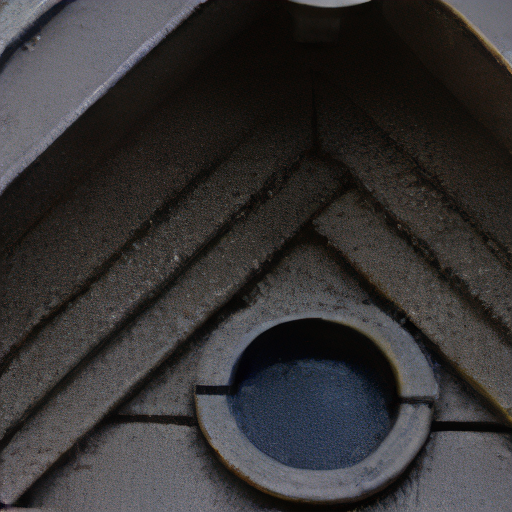}
\end{minipage}
\begin{minipage}{0.19\textwidth}
\includegraphics[width=\linewidth]{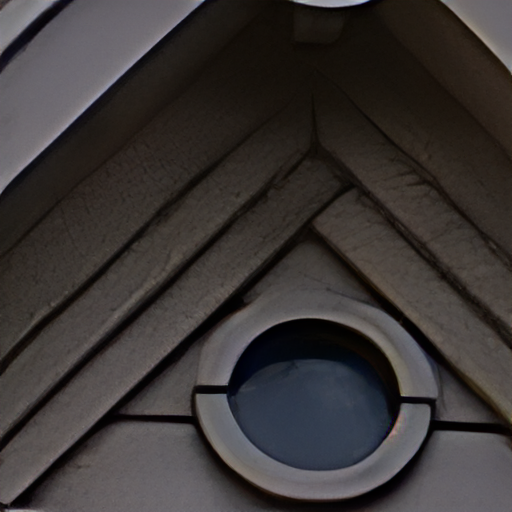}
\end{minipage}
\begin{minipage}{0.19\textwidth}
\includegraphics[width=\linewidth]{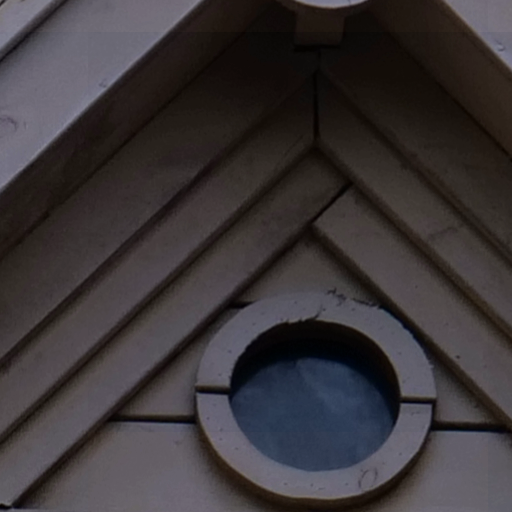}
\end{minipage}
\begin{minipage}{0.19\textwidth}
\centering
(a)
\end{minipage}
\begin{minipage}{0.19\textwidth}
\centering
(b)
\end{minipage}
\begin{minipage}{0.19\textwidth}
\centering
(c)
\end{minipage}
\begin{minipage}{0.19\textwidth}
\centering
(d)
\end{minipage}
\begin{minipage}{0.19\textwidth}
\centering
(e)
\end{minipage}
\caption{Qualitative results on samples from the RealSR and DRealSR datasets capturing real camera degradations. (a) input LR, (b) StableSR 200-steps (c) Resshift 15-Steps (d) Ours 1-step (e) ground truth. ResShift and StableSR include $173M$ and $1043M$ parameters and both require $> 2,000$ GFLOPs to process a $128\times128$ patch, resp. Ours includes only 169M parameters and requires 142 GFLOPs. Despite being hugely cost and parameter-efficient, our method works competitively with the larger baselines.}
\label{fig:sup-real}
\end{figure*}

In addition to the results presented in Figure~\ref{fig:qualititavie} of the main paper, we provide additional qualitative examples in Figures~\ref{fig:sup-synth} and \ref{fig:sup-real} showing more results on both synthetic and real datasets. These examples further support the results of Table~\ref{tab:comp-to-sd}, where we can see that our model, with 1 denoising step, \emph{consistently} outperforms ResShift~\cite{resshift_neurips23}, which is the most closely related approach to ours in terms of training paradigm, albeit it uses 15 denoising steps. In addition, in all cases we are visually at least on-par with StableSR~\cite{stablesr_arx23}, which employs a much larger model and 200 denoising steps, thereby requiring close to 15 minutes to generate each image while running on an A100 GPU. In contrast, our model, running on a mobile phone, takes $\sim$38 msec to $4\times$ upscale the same $128\times128$ images.

